\documentclass[10pt,twocolumn]{article}

\usepackage[utf8]{inputenc}
\usepackage[T1]{fontenc}
\usepackage{times}
\usepackage{graphicx}
\usepackage{amsmath,amssymb}
\usepackage{booktabs}
\usepackage{multirow}
\usepackage[table]{xcolor}
\usepackage{algorithm}
\usepackage{algpseudocode}
\usepackage{float}        
\usepackage{placeins}     
\usepackage{fancyvrb}     

\definecolor{promptbg}{gray}{0.95}
\DefineVerbatimEnvironment{promptbox}{Verbatim}{%
  fontsize=\small,
  frame=single,
  framerule=0.5pt,
  rulecolor=\color{gray},
  fillcolor=\color{promptbg},
  xleftmargin=6pt,
  xrightmargin=6pt
}

\usepackage[margin=1in]{geometry}

\usepackage[breaklinks,colorlinks,linkcolor=blue,citecolor=blue,urlcolor=blue]{hyperref}
\usepackage{cleveref}

\usepackage{microtype}

\DeclareMathOperator*{\argmin}{arg\,min}
\DeclareMathOperator*{\argmax}{arg\,max}

\title{AgenticPruner: MAC-Constrained Neural Network Compression via LLM-Driven Strategy Search}

\author{
    Shahrzad Esmat \quad
    Mahdi Banisharif \quad
    Ali Jannesari \\[0.5em]
    Iowa State University \\[0.3em]
    \texttt{\{sesmat, msharif, jannesar\}@iastate.edu}
}

\date{}

\begin{document}

\maketitle

\begin{abstract}
Neural network pruning remains essential for deploying deep learning models on resource-constrained devices, yet existing approaches primarily target parameter reduction without directly controlling computational cost. This yields unpredictable inference latency in deployment scenarios where strict Multiply-Accumulate (MAC) operation budgets must be met. We propose AgenticPruner, a framework utilizing large language models to achieve MAC-constrained optimization through iterative strategy learning. Our approach coordinates three specialized agents: a Profiling Agent that analyzes model architecture and MAC distributions, a Master Agent that orchestrates the workflow with divergence monitoring, and an Analysis Agent powered by Claude 3.5 Sonnet that learns optimal strategies from historical attempts. Through in-context learning, the Analysis Agent improves convergence success rate from 48\% to 71\% compared to grid search. Building upon isomorphic pruning's graph-based structural grouping, our method adds context-aware adaptation by analyzing patterns across pruning iterations, enabling automatic convergence to target MAC budgets within user-defined tolerance bands.

We validate our framework on ImageNet-1K across ResNet, ConvNeXt, and DeiT architectures. On CNNs, our approach achieves MAC targeting while maintaining or improving accuracy: ResNet-50 reaches 1.77G MACs with 77.04\% accuracy (+0.91\% vs baseline); ResNet-101 achieves 4.22G MACs with 78.94\% accuracy (+1.56\% vs baseline). For ConvNeXt-Small, pruning to 8.17G MACs yields 1.41$\times$ GPU and 1.07$\times$ CPU speedup with 45\% parameter reduction. On Vision Transformers, we demonstrate MAC-budget compliance within user-defined tolerance bands (typically +1\% to +5\% overshoot, $-$5\% to $-$15\% undershoot), establishing feasibility for deployment scenarios requiring strict computational guarantees.
\end{abstract}

\section{Introduction}
\label{sec:intro}

The deployment of deep neural networks under strict computational budgets demands careful optimization of both model size and computational cost. While modern architectures like ResNet~\cite{he2016deep}, ConvNeXt~\cite{liu2022convnet}, and Vision Transformers~\cite{dosovitskiy2020image,touvron2021training} achieve impressive accuracy on challenging benchmarks such as ImageNet-1K~\cite{deng2009imagenet}, their substantial computational requirements often exceed the capabilities of mobile and embedded platforms. This has driven research in neural network compression, with pruning emerging as one of the most effective techniques for reducing model complexity while preserving accuracy~\cite{cheng2023survey,he2024structured}.

Traditional pruning methods primarily focus on parameter reduction, operating under the assumption that reducing the number of weights will proportionally decrease computational cost. However, this assumption often proves unreliable in practice. Parameter count and multiply-accumulate (MAC) operations exhibit a complex, architecture-dependent relationship. For instance, pruning 50\% of parameters may reduce MACs by anywhere from 30\% to 65\%, creating significant uncertainty for deployment scenarios with strict computational budgets. This unpredictability becomes particularly problematic when targeting specific hardware constraints, where exceeding MAC limits by even a small margin can prevent deployment entirely.

Recent advances in structured pruning have introduced sophisticated approaches to identify and remove redundant components. Isomorphic pruning~\cite{fang2024isomorphic} represents a significant step forward by grouping structurally similar sub-networks using graph isomorphism and pruning within these groups. This approach ensures fair comparison between components and has demonstrated strong results across both CNNs and Vision Transformers. However, like its predecessors, isomorphic pruning remains fundamentally parameter-centric, offering limited direct control over the resulting computational budget. Moreover, these methods typically require extensive manual hyperparameter tuning to achieve desired MAC targets, with each failed attempt necessitating complete retraining.

We address these limitations by introducing a multi-agent framework that inverts the traditional pruning paradigm: rather than targeting parameter reduction and hoping for proportional MAC savings, we directly optimize for target MAC budgets within defined tolerance bands while allowing parameters to adjust as needed. Our approach coordinates three specialized agents to achieve this goal. The \textbf{Profiling Agent} analyzes model architecture and quantifies baseline MAC distributions across layer types. The \textbf{Master Agent} orchestrates the overall workflow with divergence monitoring and maintains pruning history. The \textbf{Analysis Agent}, powered by Claude 3.5 Sonnet, learns optimal pruning strategies by analyzing patterns across iterative attempts, automatically adjusting configurations to converge on target MAC budgets within user-defined tolerance bands.

Building upon the structural grouping foundation of isomorphic pruning~\cite{fang2024isomorphic}, our framework adds history-aware adaptation through large language model (LLM) reasoning~\cite{yao2022react}. The Analysis Agent examines historical pruning results, identifies patterns in successful and failed attempts, and predicts optimal configurations for subsequent iterations. This learning mechanism enables automatic convergence to MAC targets typically within 3-5 revisions, eliminating the need for manual hyperparameter search. The system supports asymmetric tolerance policies (e.g., +1\%/-5\%), reflecting real-world deployment constraints where slight underutilization is preferable to budget violations.

Our contributions are threefold: (1) We introduce the first pruning framework with direct MAC-budget optimization as the primary objective, achieving target computational costs within narrow tolerance bands; (2) We develop a multi-agent coordination architecture that combines structural pruning insights with LLM-guided strategy learning, enabling automatic iterative refinement; (3) We validate our approach on ImageNet-1K across ResNet, ConvNeXt, and DeiT architectures, demonstrating competitive accuracy with predictable computational costs and significant practical speedups.

The remainder of this paper is organized as follows. Section~\ref{sec:relatedwork} reviews related work in neural network pruning and model compression. Section~\ref{sec:methodology} details our multi-agent framework and MAC-aware pruning algorithm. Section~\ref{sec:experiments} presents experimental setup and implementation details. Section~\ref{sec:results} analyzes our results across multiple architectures, compares against state-of-the-art methods, discusses limitations and future directions. Section~\ref{sec:conclusion} concludes.

\section{Related Work}
\label{sec:relatedwork}

\subsection{Neural Network Pruning}

Neural network pruning has emerged as a fundamental technique for model compression, with extensive research spanning several decades. Early pioneering work includes Optimal Brain Damage~\cite{lecun1990optimal} and Optimal Brain Surgeon~\cite{hassibi1993optimal}, which used second-order Taylor expansions to identify parameters for removal. Modern pruning approaches can be broadly categorized into unstructured and structured pruning~\cite{cheng2023survey,he2024structured}. Unstructured pruning removes individual weights, creating sparse networks that require specialized hardware support~\cite{han2015learning}. In contrast, structured pruning removes entire channels, filters, or attention heads, producing models that are immediately compatible with standard hardware accelerators~\cite{li2017pruning,he2017channel}.

\subsection{Importance Criteria for Pruning}

Determining which parameters to prune requires effective importance criteria. Magnitude-based pruning~\cite{li2017pruning} ranks filters by their L1 or L2 norms, operating under the assumption that smaller weights contribute less to network output. Taylor expansion-based methods~\cite{molchanov2019importance} approximate the change in loss induced by removing each parameter, providing a more principled approach that has demonstrated superior performance across diverse architectures. First-order Taylor importance estimates the impact as $|\mathcal{L} - \mathcal{L}_{\theta=0}| \approx |g \cdot \theta|$ where $g$ represents gradients and $\theta$ denotes parameters. Hessian-based approaches~\cite{yang2023nvit} incorporate second-order information but require additional computational overhead. While these criteria excel at ranking parameters within a single layer, they struggle with cross-layer dependencies inherent in structured pruning.

\subsection{Graph-Based Dependency Modeling}

Structural pruning presents unique challenges due to parameter coupling across layers. Removing an output channel from one convolutional layer necessitates removing the corresponding input channel in subsequent layers. DepGraph~\cite{fang2023depgraph} introduced an automated framework for modeling these dependencies by representing neural networks as computational graphs. The approach builds a dependency graph $G(V,E)$ where vertices represent operations and edges encode parameter dependencies. DepGraph enables any-architecture pruning by automatically identifying coupled parameters that must be pruned simultaneously, eliminating the need for manual dependency analysis for each architecture type. This work demonstrated state-of-the-art results across CNNs, RNNs, GNNs, and Transformers, establishing graph-based modeling as a cornerstone for modern structured pruning.

\subsection{Isomorphic Pruning for Vision Models}

Building upon DepGraph's dependency modeling, Isomorphic Pruning~\cite{fang2024isomorphic} addressed the critical problem of fair comparison between structurally heterogeneous components. The key insight is that network sub-structures with identical computational topologies (graph isomorphisms) tend to exhibit similar importance distributions, making them directly comparable for ranking purposes. The method groups sub-structures by their graph representations and applies importance ranking within each isomorphic group rather than globally. This approach prevents the unfair comparison of, for example, attention blocks against MLP blocks, which operate at vastly different computational scales. Isomorphic Pruning demonstrated superior accuracy retention compared to global pruning baselines on both CNNs (ResNet, ConvNeXt) and Vision Transformers (DeiT), achieving state-of-the-art results on ImageNet-1K. However, like prior work, it primarily targets parameter reduction with indirect control over computational cost.

\subsection{Vision Transformer Pruning}

The success of Vision Transformers~\cite{dosovitskiy2020image,touvron2021training} has motivated specialized pruning techniques. Early approaches~\cite{zhu2021vision} introduced dimension-wise sparsity to identify important dimensions for pruning. WDPruning~\cite{yu2022width} simultaneously reduces both width and depth, introducing learnable parameters to adapt transformer dimensions and shallow classifiers for depth pruning. NViT~\cite{yang2023nvit} employs Hessian-aware saliency for global structural pruning, achieving significant speedups with minimal accuracy loss. SAViT~\cite{zheng2022savit} introduces joint importance that captures structural interactions between heterogeneous components (attention heads, MLP blocks), using Taylor-based approximations for balanced pruning. These methods have demonstrated that Vision Transformers contain substantial redundancy, particularly in attention mechanisms and deeper layers, enabling aggressive compression while maintaining competitive accuracy.

\subsection{MAC and FLOPs Optimization}

While most pruning methods target parameter reduction, a limited body of work has explored direct control over multiply-accumulate operations (MACs) or floating-point operations (FLOPs). A natural question arises: \textit{can simple threshold-based methods achieve precise MAC targets?} Global magnitude pruning~\cite{blalock2020state} ranks all parameters by magnitude and removes the smallest until a target sparsity is reached. While surprisingly effective for unstructured pruning~\cite{frankle2018lottery}, this approach struggles with structured pruning under MAC constraints for three reasons: (1)~heterogeneous layers have vastly different MAC contributions but similar magnitude distributions, causing global thresholds to over-prune some layers while under-pruning others; (2)~structural coupling means removing channels affects downstream layers in architecture-dependent ways; and (3)~the MAC-to-accuracy relationship is non-monotonic, where small ratio changes can cause large jumps due to layer-wise rounding effects.

Two methods most directly address computational budget constraints. FALCON~\cite{meng2024falcon} formulates FLOPs-constrained pruning as integer linear programming (ILP), solving for optimal layer-wise sparsity in a single optimization pass. While mathematically elegant, ILP treats pruning as a one-shot decision: if the resulting model fails accuracy requirements, the entire optimization must restart with adjusted constraints, discarding information about why the configuration failed. HALP~\cite{shen2021halp} incorporates hardware-aware latency predictors to guide pruning toward configurations that achieve target inference speeds on specific devices. However, HALP optimizes for latency rather than MAC budgets, and like FALCON, provides no mechanism for learning from unsuccessful attempts.

Our approach addresses this gap through iterative strategy learning. Rather than treating each pruning attempt as independent, the Analysis Agent maintains a history of configurations, their achieved MACs, and accuracy outcomes. When a configuration overshoots or undershoots the target, the agent analyzes \textit{why}---identifying patterns such as ``aggressive early-layer pruning caused MAC undershoot'' or ``conservative MLP ratios left accuracy headroom''---and predicts corrective adjustments for the next iteration. This learning mechanism reduces iterations from 68 (fixed heuristics) to 31 while improving success rate from 48\% to 71\% compared to grid search (Table~\ref{tab:app_agent_ablation}). The key distinction is \textit{adaptive refinement}: FALCON and HALP optimize once, while our method learns across attempts.

\subsection{LLM-Based Optimization}

Large language models have recently been explored for neural architecture search and optimization tasks, raising the question of whether our LLM-guided pruning is merely ``hyperparameter tuning with extra steps.'' We argue it is fundamentally different. LLMatic~\cite{nasir2024llmatic} combines LLM code generation with quality-diversity algorithms for NAS, generating novel architecture code that must be trained from scratch, with each candidate requiring full training (hours to days). AutoML-GPT~\cite{zhang2024automl} tunes training hyperparameters (learning rate, batch size) where the search space is well-understood and candidates can be evaluated quickly. Note that LLM-Pruner~\cite{ma2023llmpruner} addresses a distinct problem: pruning \textit{of} large language models using gradient-based importance, rather than using LLMs \textit{for} pruning decisions as we do.

Our setting differs in three crucial ways. First, \textit{the search space is architecture-dependent}: pruning ratios interact with structural constraints (isomorphic groups, channel dependencies) that vary across ResNet, ConvNeXt, and DeiT. Second, \textit{feedback is delayed and noisy}: a pruning configuration's quality is only known after fine-tuning (minutes to hours), and small ratio changes can cause catastrophic accuracy collapse. Third, \textit{the objective is multi-dimensional}: we simultaneously target MAC budget compliance AND accuracy retention, with complex trade-offs that simple grid search cannot efficiently navigate.

The Analysis Agent addresses these challenges through pattern recognition across heterogeneous structural groups: it learns that ``MLP blocks tolerate 1.5$\times$ higher pruning ratios than QKV projections in DeiT'' or ``ResNet's later stages contribute 60\% of MACs but only 20\% to accuracy.'' This architectural reasoning, grounded in pruning history, enables convergence in 3-5 iterations where fixed heuristics require 68+ attempts (Table~\ref{tab:app_agent_ablation}).

Our LLM-based approach offers distinct advantages over alternative optimization paradigms. Reinforcement learning methods such as AMC~\cite{he2018amc} require hundreds of pruning episodes to learn effective policies, whereas our Analysis Agent converges in 3--5 revisions through few-shot learning from pruning history. Bayesian Optimization treats configurations as numerical vectors without semantic understanding of architectural constraints (e.g., ``attention heads must remain divisible by 8'' or ``early convolutional layers are sensitivity-critical''). In contrast, our Analysis Agent reasons explicitly about structural dependencies through natural language, enabling architecture-aware decisions that black-box optimizers cannot express. Appendix~\ref{app:reasoning_trace} provides an example reasoning trace demonstrating this semantic understanding.

\subsection{Contribution}

Our approach differs fundamentally from prior work in three key aspects. First, we invert the traditional pruning paradigm by prioritizing MAC-budget optimization over parameter reduction, ensuring predictable computational costs. Second, we introduce a multi-agent architecture where specialized agents coordinate pruning decisions: a Profiling Agent analyzes model structure and MAC distributions, a Master Agent orchestrates workflow and detects failures, and an Analysis Agent powered by LLMs learns optimal pruning strategies from historical attempts. Third, we enable automatic iterative refinement that converges to target MAC budgets within user-defined tolerance bands, typically in 3-5 revisions, removing the need for manual tuning. While building upon the structural grouping insights of isomorphic pruning~\cite{fang2024isomorphic}, we augment it with history-aware adaptation through LLM-guided learning, achieving MAC-constrained optimization unavailable in prior methods.

\section{Methodology}
\label{sec:methodology}

AgenticPruner is a multi-agent framework for MAC-constrained neural network pruning that achieves strict computational budget control through coordinated decision-making and LLM-driven strategy learning. Unlike conventional parameter-centric methods, our approach directly optimizes for target MAC operations, ensuring predictable deployment costs while maintaining competitive accuracy.

\subsection{Problem Formulation}

Given a pre-trained neural network $\mathcal{M}$ with baseline MACs $M_{\text{base}}$ and a target MAC budget $M_{\text{target}}$, our objective is to find a pruned network $\mathcal{M}'$ satisfying:
\begin{equation}
    M_{\text{target}} \times (1 - \delta_{\text{under}}) \leq M(\mathcal{M}') \leq M_{\text{target}} \times (1 + \delta_{\text{over}})
\end{equation}
where $M(\cdot)$ measures multiply-accumulate operations, and $\delta_{\text{over}}, \delta_{\text{under}}$ define asymmetric tolerance bands (typically $\delta_{\text{over}} = 0.01$, $\delta_{\text{under}} = 0.05$). This formulation reflects real-world deployment constraints where exceeding computational budgets may violate hardware limitations, while modest underutilization is acceptable.

For Vision Transformers, we use isomorphic grouping~\cite{fang2024isomorphic} to partition the network into structurally equivalent sub-networks $\{\mathcal{G}_1, \mathcal{G}_2, \ldots, \mathcal{G}_K\}$ (e.g., MLP blocks, QKV projections). Each group $\mathcal{G}_k$ receives a target MAC allocation $M_k$ such that $\sum_{k=1}^K M_k \approx M_{\text{target}}$. For CNNs, we employ channel-wise pruning with layer-dependent ratios optimized to meet the global MAC constraint.

\subsection{Multi-Agent Architecture}

\begin{figure*}[t]
  \centering
  \includegraphics[width=\textwidth]{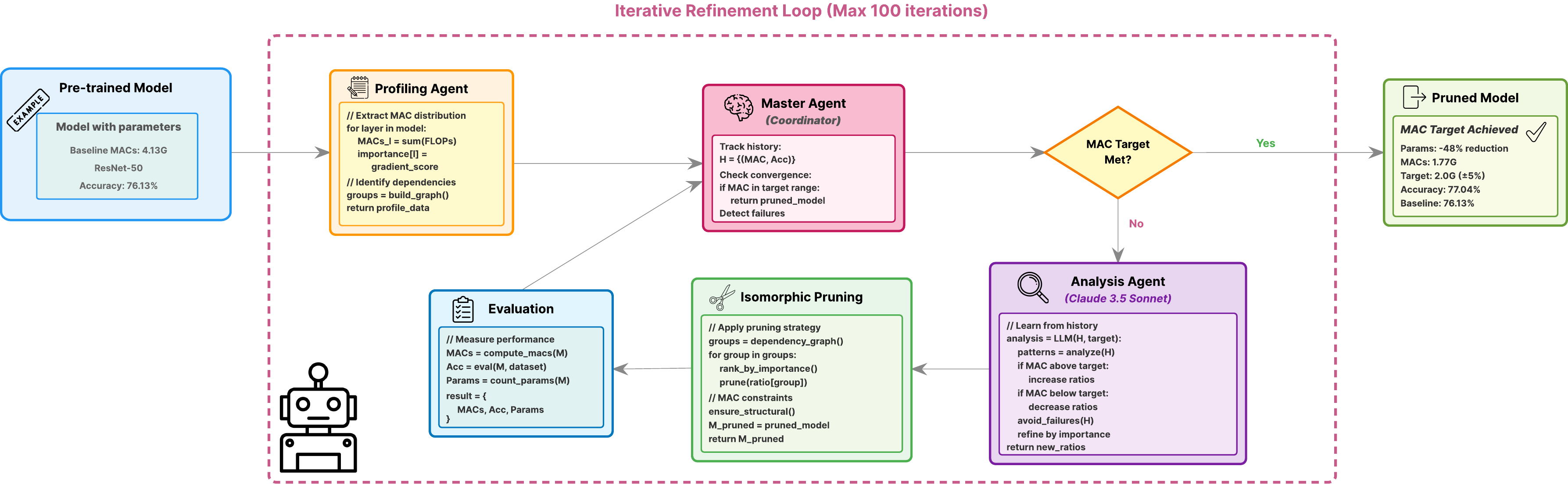}
  \caption{\textbf{Multi-agent framework architecture.} The Profiling Agent analyzes model structure and baseline MAC distribution. The Master Agent orchestrates workflow and detects convergence patterns. The Analysis Agent, powered by an LLM, learns optimal pruning strategies from historical attempts. Iterative refinement continues until MAC targets are met within tolerance.}
  \label{fig:architecture}
\end{figure*}

Our framework comprises three specialized agents that coordinate to achieve MAC-budget objectives (Figure~\ref{fig:architecture}). Each agent has distinct responsibilities and operates with specific domain knowledge, enabling systematic exploration of the pruning configuration space.

\subsubsection{Profiling Agent}
\label{sec:profiling}

The Profiling Agent characterizes the computational landscape of the input model through static analysis and forward-pass profiling. Unlike simple PyTorch hooks that only count operations, our agent leverages DepGraph~\cite{fang2023depgraph} to construct a full dependency graph, enabling accurate MAC prediction \textit{after} pruning by tracking how channel removal propagates through coupled layers. For each layer $\ell$, it computes MAC contributions using operator-specific formulas. For convolutional layers: $M_{\ell} = H_{\text{out}} \times W_{\text{out}} \times C_{\text{out}} \times C_{\text{in}} \times K_H \times K_W$, and for linear layers: $M_{\ell} = D_{\text{in}} \times D_{\text{out}}$, where $H, W$ denote spatial dimensions, $C$ channels, $K$ kernel sizes, and $D$ hidden dimensions. Note that we target theoretical MACs rather than hardware latency, as MAC counts provide architecture-agnostic deployment guarantees; device-specific latency optimization remains future work (Section~\ref{sec:limitations}).

The agent identifies architectural dependencies through computational graph traversal, detecting coupled layers that must maintain dimensional compatibility. For Vision Transformers, it partitions layers into isomorphic groups based on computational topology~\cite{fang2024isomorphic}. The profiling output includes: (1) baseline MAC distribution $\{M_{\ell}\}_{\ell=1}^L$ across all layers, (2) structural constraints preventing accuracy collapse, (3) isomorphic groupings for fair importance comparison, and (4) dataset-specific considerations (e.g., ImageNet requires more conservative pruning than CIFAR-10 due to task complexity).

\subsubsection{Master Agent}
\label{sec:master}

The Master Agent serves as the central coordinator, maintaining global state across pruning iterations and implementing strategic oversight. At each revision $r$, it analyzes the pruning history $\mathcal{H} = \{(\mathbf{s}_i, M_i, A_i)\}_{i=0}^{r-1}$, where $\mathbf{s}_i$ denotes the strategy configuration, $M_i$ the achieved MACs, and $A_i$ the resulting accuracy.

The Master Agent detects convergence patterns and dangerous configuration zones through statistical analysis of historical outcomes. It identifies parameter ranges that consistently lead to failures (e.g., MAC overshoot or catastrophic accuracy collapse) and provides explicit avoidance guidance to the Analysis Agent. When multiple attempts achieve similar MAC targets, the Master Agent selects the configuration with highest accuracy and signals termination.

The Master Agent implements four coordinated mechanisms to guide convergence. First, it monitors progress by tracking the MAC error reduction rate $\Delta_r = |M_r - M_{\text{target}}| - |M_{r-1} - M_{\text{target}}|$, flagging stagnation when improvements plateau. Second, it detects accuracy collapse (drops exceeding 10\% from baseline) and triggers conservative fallback strategies. Third, pattern recognition groups historical attempts by MAC error magnitude, learning empirical scaling relationships between configuration adjustments and achieved MACs. Finally, early stopping terminates the search when three consecutive revisions achieve MAC targets within tolerance with minimal accuracy variance.

\subsubsection{Analysis Agent}
\label{sec:analysis}

The Analysis Agent uses a large language model (Claude 3.5 Sonnet) to determine optimal pruning strategies by reasoning about historical patterns. This is \textit{in-context learning}, not fine-tuning: the complete pruning history is included in each prompt, leveraging the LLM's long context window to maintain sensitivity patterns across iterations without gradient updates. Given the current state and Master Agent guidance, the LLM generates configuration proposals through structured prompting.

The agent's core function implements:
\begin{equation}
    \mathbf{s}_r = \text{LLM}(\mathcal{H}, M_{\text{target}}, \mathcal{P}_{\text{profiling}}, \mathcal{G}_{\text{master}})
\end{equation}
where $\mathcal{P}_{\text{profiling}}$ contains architectural insights and $\mathcal{G}_{\text{master}}$ provides strategic guidance. The LLM prompt includes: (1) complete pruning history with MAC errors and accuracy outcomes, (2) target MAC budget and tolerance bounds, (3) architectural characteristics and baseline MAC distribution, (4) identified failure zones and recommended avoidance strategies.

For Vision Transformers, the LLM determines group-wise multipliers:
\begin{equation}
    \mathbf{s}_{\text{ViT}} = \{m_{\text{mlp}}, m_{\text{qkv}}, m_{\text{proj}}, m_{\text{head}}\}
\end{equation}
where each multiplier scales the base pruning ratio to achieve group-specific MAC allocations. For CNNs, it outputs layer-dependent channel pruning ratios, importance criterion selection (Taylor, magnitude, or random), and rounding granularity for hardware efficiency.

The LLM's reasoning capabilities enable several key advantages: (1) \textbf{pattern generalization} across architectures and datasets, (2) \textbf{failure avoidance} by learning from unsuccessful attempts, (3) \textbf{sensitivity estimation} through observed MAC-configuration relationships, and (4) \textbf{strategy adaptation} based on convergence trends. Unlike fixed heuristics or learned policies requiring extensive training data, the LLM operates in a few-shot regime, adapting from just 2-3 prior attempts. See Appendix~\ref{app:reasoning_trace} for an example reasoning trace demonstrating the LLM's semantic understanding of architectural constraints.

\subsection{MAC-Budget Optimization Algorithm}

\begin{figure}[t]
\centering
\includegraphics[width=0.95\linewidth]{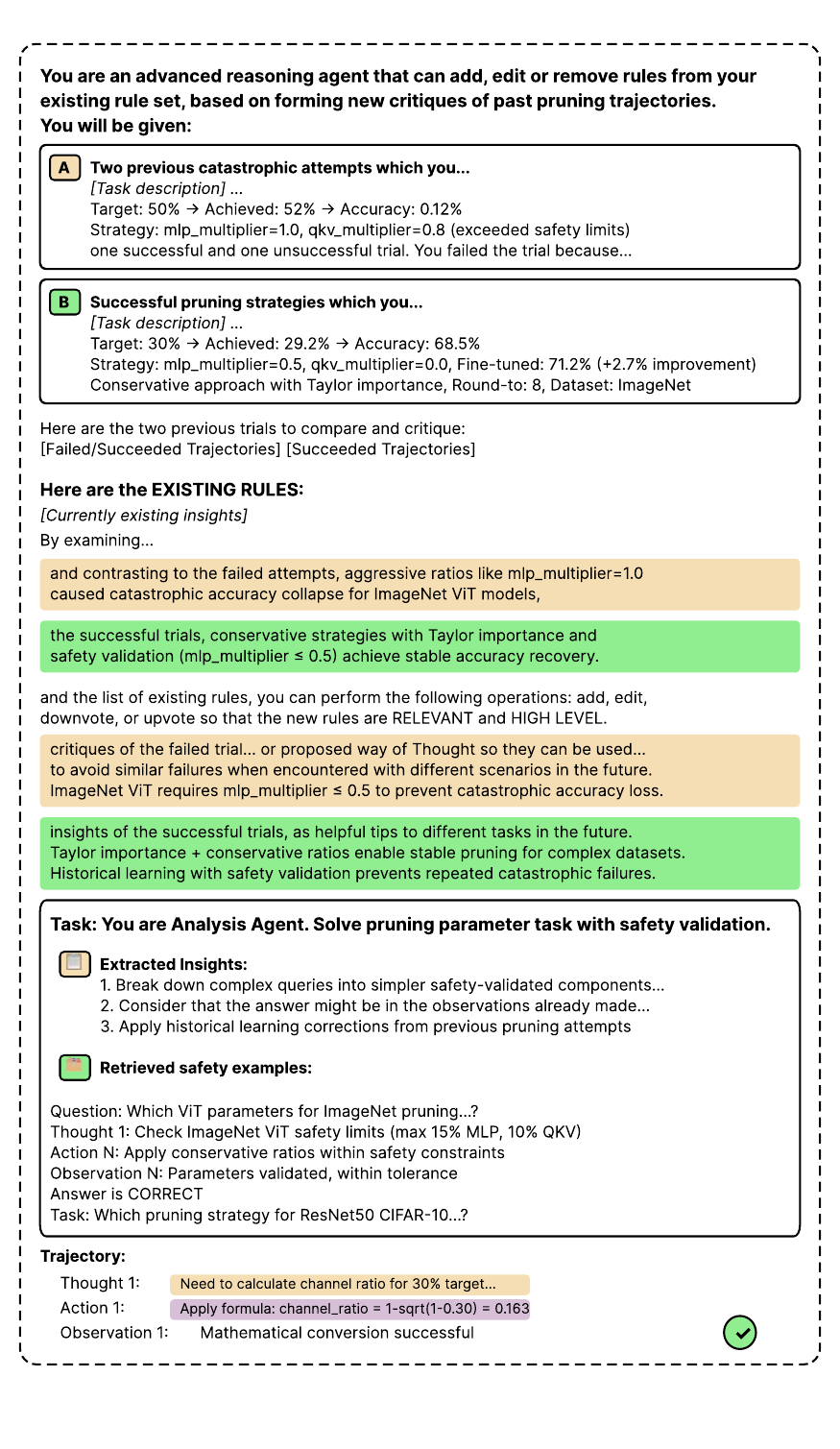}
\caption{\textbf{Iterative MAC-budget optimization workflow.} Each revision refines the pruning strategy based on LLM analysis of previous attempts until MAC targets are achieved. The workflow typically converges within 3-5 revisions.}
\label{fig:workflow}
\end{figure}

Algorithm~\ref{alg:macaware} presents our complete optimization procedure. The workflow alternates between strategy generation (Analysis Agent), pruning execution, and convergence assessment (Master Agent) until MAC targets are satisfied.

\begin{algorithm}[t]
\caption{MAC-Aware Multi-Agent Pruning}
\label{alg:macaware}
\small
\begin{algorithmic}[1]
\Require Pre-trained model $\mathcal{M}$, target MACs $M_{\text{target}}$, tolerances $\delta_{\text{over}}, \delta_{\text{under}}$, max revisions $R_{\text{max}}$
\Ensure Pruned model $\mathcal{M}'$ with $M(\mathcal{M}') \in [M_{\text{target}}(1-\delta_{\text{under}}), M_{\text{target}}(1+\delta_{\text{over}})]$
\State $\mathcal{P} \gets$ \textsc{ProfilingAgent}($\mathcal{M}$)
\State $\mathcal{H} \gets \emptyset$; $\mathcal{C} \gets \emptyset$ \Comment{History and candidates}
\For{$r = 0$ to $R_{\text{max}} - 1$}
    \State $\mathcal{G} \gets$ \textsc{MasterAgent}($\mathcal{H}, M_{\text{target}}$)
    \State $\mathbf{s}_r \gets$ \textsc{AnalysisAgent}($\mathcal{H}, \mathcal{P}, \mathcal{G}, M_{\text{target}}$)
    \State $\mathcal{M}_r \gets$ \textsc{Prune}($\mathcal{M}, \mathbf{s}_r$) \Comment{Always from original weights}
    \State $M_r \gets$ \textsc{CountMACs}($\mathcal{M}_r$)
    \If{$M_{\text{target}}(1-\delta_{\text{under}}) \leq M_r \leq M_{\text{target}}(1+\delta_{\text{over}})$}
        \State $\mathcal{M}_r^{\text{ft}} \gets$ \textsc{InlineFineTune}($\mathcal{M}_r$) \Comment{5 epochs on 30\% data}
        \State $A_r \gets$ \textsc{Eval}($\mathcal{M}_r^{\text{ft}}$)
        \State $\mathcal{C} \gets \mathcal{C} \cup \{(\mathcal{M}_r^{\text{ft}}, M_r, A_r)\}$
        \If{\textsc{MasterAgent.ShouldStop}($\mathcal{H}, \mathcal{C}$)}
            \State \Return \textsc{ExtendedFineTune}($\argmax_\mathcal{C} A$) \Comment{Full training}
        \EndIf
    \EndIf
    \State $\mathcal{H} \gets \mathcal{H} \cup \{(\mathbf{s}_r, M_r)\}$
\EndFor
\State \Return \textsc{ExtendedFineTune}($\argmax_\mathcal{C} A$) \Comment{Best candidate}
\end{algorithmic}
\end{algorithm}

\textbf{Search independence.} Each pruning iteration operates on the original pre-trained model $\mathcal{M}$ to maintain a consistent search space. Line 6 always prunes from the same baseline weights, not from previously fine-tuned models. This design allows the Analysis Agent to learn stable MAC-to-configuration relationships without compounding effects from iterative weight updates. Inline fine-tuning (Line 9) produces $\mathcal{M}_r^{\text{ft}}$ for accuracy evaluation only; the next iteration's pruning still operates on the original $\mathcal{M}$. Only after convergence is the best candidate subjected to extended fine-tuning for deployment.

\textbf{Pruning execution.} Given strategy $\mathbf{s}_r$, we employ group-wise pruning based on Taylor importance~\cite{molchanov2019importance}. For each isomorphic group $\mathcal{G}_k$, we compute first-order Taylor approximation scores:
\begin{equation}
    \mathcal{I}(\theta_i) = \left| \frac{\partial \mathcal{L}}{\partial \theta_i} \cdot \theta_i \right|
\end{equation}
where $\theta_i$ represents the $i$-th structural unit (channel, head, etc.) and $\mathcal{L}$ is the loss on a calibration set. Units are ranked within each group independently, and the bottom fraction determined by $\mathbf{s}_r$ are removed while preserving structural dependencies.

\textbf{Iterative refinement.} The key innovation lies in systematic adjustment across revisions. If revision $r$ achieves $M_r > M_{\text{target}}$ (overshoot), the Analysis Agent increases pruning aggressiveness. Conversely, undershoot ($M_r < M_{\text{target}}$) triggers more conservative configurations. The LLM learns empirical sensitivity relationships (e.g., increasing MLP multiplier by 0.1 reduces MACs by approximately 0.3G for DeiT-Tiny) and applies proportional corrections.

\subsection{Safety Validation and Constraints}

To prevent catastrophic model degradation, we enforce four architectural integrity constraints before executing each pruning attempt. Dimension compatibility ensures layer input/output dimensions remain valid after pruning: for example, attention dimensions must remain divisible by the number of heads. Minimal capacity preservation prohibits pruning ratios exceeding 85\% for critical components such as QKV projections, preventing complete information loss. Rounding granularity aligns channel counts to hardware-friendly multiples (typically 8 or 16) for efficient SIMD execution on target devices. Finally, dependency preservation maintains coupled dimensions across connected layers through graph-based analysis~\cite{fang2023depgraph}, ensuring that pruning one layer does not invalidate dependent computations downstream. If a proposed strategy violates any constraint, the Analysis Agent regenerates configurations with stricter bounds until valid.

\subsection{Implementation Details}

We implement our framework in PyTorch~\cite{paszke2019pytorch} building on the \texttt{torch\_pruning} library~\cite{fang2023depgraph} for dependency-aware pruning operations. The LLM interface uses Claude 3.5 Sonnet via OpenRouter API with temperature 0 for deterministic strategy generation. Importance scores are computed using Taylor approximation on 512 calibration samples from the training set.

For Vision Transformers, we prune MLP blocks, QKV projections, and attention heads within each isomorphic group. For CNNs, we apply structured channel pruning layer-by-layer. Hardware rounding typically uses multiples of 8 (matching SIMD vector widths), though this is configurable per deployment target.

The complete pruning workflow (Algorithm~\ref{alg:macaware}) typically converges within 3-5 revisions for most architectures. We employ a two-stage fine-tuning strategy for computational efficiency: during search, only candidates within MAC tolerance receive brief \textit{inline fine-tuning} (5 epochs on a 30\% data subset) to assess recovery potential; after convergence, the best candidate undergoes \textit{extended fine-tuning} (up to 300 epochs on the full dataset) for final accuracy recovery. Models outside MAC tolerance are discarded without fine-tuning, making the search process efficient.

\section{Experiments}
\label{sec:experiments}

We evaluate AgenticPruner across diverse architectures and compare against state-of-the-art structured pruning methods. Our experiments demonstrate MAC-budget compliance with competitive accuracy retention and practical speedups on both GPU and CPU devices.

\subsection{Experimental Setup}

\textbf{Dataset.} We conduct all experiments on ImageNet-1K~\cite{deng2009imagenet}, which contains 1.28 million training images and 50,000 validation images across 1,000 object categories. To accelerate the iterative pruning search, we use a stratified 30\% random subset of the training data (384K images) during the multi-agent optimization phase; ablation studies in Appendix~D use a further reduced 10\% subset (128K images) for rapid hyperparameter exploration. Importance scores and zero-shot evaluations are computed on the respective subsets. Final fine-tuning and all reported accuracy numbers in Tables~1--3 use the complete training set. Images are resized to 256$\times$256 and center-cropped to 224$\times$224 for training and evaluation. We apply standard data augmentation including random horizontal flipping, color jittering, and AutoAugment~\cite{cubuk2019autoaugment} following best practices for training compressed models.

\textbf{Architectures.} We evaluate on three architecture families representing diverse computational patterns:
\begin{itemize}
    \item \textbf{ResNet}~\cite{he2016deep}: ResNet-50 (25.6M parameters, 4.1G MACs) and ResNet-101 (44.6M parameters, 7.8G MACs). These residual networks employ bottleneck blocks with $3\times3$ convolutions and skip connections.
    \item \textbf{ConvNeXt}~\cite{liu2022convnet}: ConvNeXt-Tiny (28.6M parameters, 4.5G MACs), ConvNeXt-Small (50.2M parameters, 8.7G MACs), and ConvNeXt-Base (88.6M parameters, 15.4G MACs). These modern CNN architectures incorporate depthwise convolutions, larger kernels ($7\times7$), and LayerNorm.
    \item \textbf{DeiT}~\cite{touvron2021training}: DeiT-Tiny (5.9M parameters, 1.3G MACs) and DeiT-Base (86.6M parameters, 17.6G MACs). These Vision Transformers use multi-head self-attention and position embeddings without convolutional layers.
\end{itemize}
All models are initialized from publicly available pre-trained checkpoints on ImageNet-1K to ensure fair comparison with prior work.

\textbf{Pruning targets.} We define MAC reduction targets ranging from 40\% to 75\% depending on baseline model capacity. For each architecture, we aim to match computational budgets of smaller official variants (e.g., pruning ConvNeXt-Base to ConvNeXt-Small MACs: 8.2G) while potentially exceeding their accuracy. We enforce asymmetric tolerance bands with stricter overshoot limits ($+0.1\%$ to $+5\%$) to meet deployment constraints and more lenient undershoot limits ($-5\%$ to $-20\%$) since modest underutilization is acceptable. Tolerances are tuned per architecture based on structural regularity and pruning sensitivity: ResNet-50 allows larger overshoot ($+5\%$) due to its regular bottleneck structure enabling predictable MAC estimation, while ConvNeXt-B uses tighter bounds ($+0.1\%$/$-11.56\%$) due to depthwise convolution sensitivity that makes precise MAC control more difficult.

\subsection{Implementation Details}

\textbf{Pruning configuration.} We implement our multi-agent framework using PyTorch 2.0~\cite{paszke2019pytorch} and the \texttt{torch\_pruning} library~\cite{fang2023depgraph} for dependency-aware structured pruning. The Analysis Agent uses Claude 3.5 Sonnet via OpenRouter API with temperature 0 for deterministic strategy generation. Importance scores are computed using Taylor approximation~\cite{molchanov2019importance} with gradients accumulated over 100 mini-batches (64 images each) following~\cite{fang2024isomorphic}.

For Vision Transformers, we prune within isomorphic groups~\cite{fang2024isomorphic}: MLP blocks, QKV projections, output projections, and attention heads. For CNNs, we apply layer-wise channel pruning with group-dependent ratios. Hardware rounding aligns channel counts to multiples of 8 for efficient SIMD execution on target devices.

The iterative optimization (Algorithm~\ref{alg:macaware}) runs for a maximum of 1000 revisions, though most configurations converge within 3-5 attempts. Candidates outside MAC tolerance require only 1-2 minutes for profiling and strategy generation; candidates within tolerance additionally undergo inline fine-tuning (5 epochs on 30\% of ImageNet), adding approximately 15-20 minutes per valid candidate on an NVIDIA H200 GPU.

\textbf{Search cost analysis.} Our LLM-guided search incurs minimal computational overhead relative to final retraining. With typical convergence in 5 iterations, each requiring 5 epochs on 30\% of ImageNet, the total search cost equals $5 \times 5 \times 0.3 = 7.5$ effective full-dataset epochs. This is negligible compared to final fine-tuning (100 epochs for CNNs, 300 for ViTs) and dramatically more efficient than reinforcement learning approaches like AMC~\cite{he2018amc}, which require 50--100 episodes with full model evaluation per episode, or neural architecture search methods requiring thousands of candidate evaluations. The search phase accounts for $<$8\% of total training time while eliminating manual hyperparameter tuning.

\textbf{Fine-tuning procedure.} After pruning to target MACs, we fine-tune compressed models to recover accuracy on a single NVIDIA H200 GPU with 128GB memory. Following~\cite{fang2024isomorphic}, we adhere to the original training protocol for each architecture family to ensure fair comparison with prior work.

\textit{ResNet and ConvNeXt (CNNs).} We train for 100 epochs using SGD with momentum 0.9, initial learning rate 0.08, and weight decay $10^{-4}$. The learning rate decays by factor 10 at epochs 30, 60, and 90 (step schedule). We use batch size 1024 without advanced augmentation techniques to match standard CNN training protocols~\cite{he2016deep,liu2022convnet}.

\textit{DeiT (Vision Transformers).} We train for 300 epochs using AdamW~\cite{loshchilov2017decoupled} with initial learning rate $5\times10^{-4}$, weight decay 0.05, and gradient clipping at 5.0. The learning rate follows a cosine annealing schedule~\cite{loshchilov2016sgdr} with 5-epoch linear warmup. We use batch size 2048 with extensive regularization: label smoothing (0.1), mixup ($\alpha=0.2$), cutmix ($\alpha=1.0$), random erasing (0.25), dropout (0.0), and stochastic depth (0.1), following DeiT training recipes~\cite{touvron2021training}.

\textit{Knowledge distillation.} For all architectures, knowledge distillation~\cite{hinton2015distilling} transfers information from the unpruned teacher to the pruned student using temperature $\tau=4.0$ and interpolation weight $\alpha_{\text{KD}}=0.3$. Model EMA~\cite{polyak1992acceleration} with decay 0.99998 provides stable evaluation metrics. Early stopping (patience of 10 epochs) monitors validation accuracy; in practice, CNNs converge within 80--90 epochs and ViTs within 200--250 epochs.

\subsection{Baseline Methods}

We compare against established structured pruning techniques spanning importance-based, geometric, and automated approaches for both CNNs and Vision Transformers.

\textbf{CNN baselines} include importance-based methods such as Taylor pruning~\cite{molchanov2019importance} with first-order gradient approximation and Group Fisher Pruning~\cite{liu2021group} using Fisher information. Geometric approaches include FPGM~\cite{he2019filter} which removes filters via geometric median, and SFP~\cite{he2018soft} with iterative soft regularization. We also compare against sparsity-inducing methods (ISTA~\cite{ye2018rethinking}), collaborative approaches that model cross-layer dependencies (CCP~\cite{peng2019collaborative}), and automated pruning methods including AutoSlim~\cite{yu2019autoslim} for one-shot architecture search and DepGraph~\cite{fang2023depgraph} for graph-based dependency modeling.

\textbf{Vision Transformer baselines} span Hessian-aware methods (NViT~\cite{yang2023nvit}), dimension-wise pruning (WDPruning~\cite{yu2022width}, ViT-Slim~\cite{chavan2022vision}), structure-aware optimization (SAViT~\cite{zheng2022savit}), explainability-guided approaches (X-Pruner~\cite{yu2023xpruner}), unified frameworks (UVC~\cite{tu2022unified}, UP-DeiT~\cite{tang2022unified}), head selection methods (GOHSP~\cite{yin2022global}), and coupled pruning (CP-ViT~\cite{yu2022cp}).

Our primary comparison is \textbf{Isomorphic Pruning}~\cite{fang2024isomorphic}, which groups parameters by computational topology for fair importance ranking but lacks MAC-budget optimization. We report their published ImageNet-1K results and additionally run experiments with identical fine-tuning for controlled comparison. All other baseline results are sourced from original publications; for methods without reported results on our target configurations, we implement using official code with recommended hyperparameters.

\subsection{Evaluation Metrics}

\textbf{Accuracy.} We report Top-1 accuracy (\%) on the ImageNet-1K validation set. Zero-shot accuracy refers to evaluation immediately after pruning without fine-tuning, while final accuracy is measured after the extended fine-tuning procedure (100 epochs for CNNs, 300 epochs for ViTs).

\textbf{MAC operations.} We measure multiply-accumulate operations (MACs) in giga-operations (G) using the \texttt{torch\_pruning} profiler, which computes exact counts via computational graph analysis. This metric directly quantifies inference computational cost and correlates with energy consumption.

\textbf{Parameters.} We report the number of trainable parameters in millions (M) after pruning. While our framework optimizes for MACs rather than parameters, parameter reduction provides additional storage benefits.

\textbf{Inference latency.} We measure wall-clock inference time on two hardware platforms:
\begin{itemize}
    \item \textbf{GPU}: NVIDIA H200 with batch size 256, reporting mean latency over 100 runs with warm-up.
    \item \textbf{CPU}: Intel Xeon Platinum 8380 (2.30GHz) with batch size 8, averaged over 100 runs.
\end{itemize}
Models are compiled with PyTorch 2.0's \texttt{torch.compile} backend for optimized execution. Latency measurements use FP32 precision and include all preprocessing overhead.

\textbf{Model size.} We report serialized model size in megabytes (MB) after pruning, which impacts storage and transmission costs in deployment scenarios.

\textbf{Memory footprint.} Peak GPU memory consumption (MB) during inference is measured via PyTorch profiler at the specified batch sizes, indicating device memory requirements.

\subsection{Ablation Studies}

To validate our design choices, we conduct ablation experiments examining:

\textbf{Agent contribution.} We evaluate simplified variants removing individual agents: (1) no Master Agent (Analysis Agent generates strategies without historical pattern guidance), (2) no Analysis Agent (fixed heuristic strategies without LLM reasoning), and (3) single-agent baseline (combined profiling and pruning without iterative refinement).

\textbf{LLM model selection.} We compare Claude 3.5 Sonnet against alternative language models (GPT-4, Llama-3-70B) to assess the impact of reasoning quality on convergence speed and final MAC accuracy.

\textbf{Importance criterion.} We test Taylor approximation versus magnitude-based and random importance scoring to quantify the contribution of gradient-informed pruning decisions.

\textbf{Tolerance bands.} We vary asymmetric tolerance settings (symmetric $\pm1\%$, symmetric $\pm5\%$, and our asymmetric $+1\%/-5\%$) to analyze the trade-off between strict MAC control and search efficiency.

\textbf{Revision budget.} We examine convergence behavior with different maximum revision limits (1, 3, 5, 10) to determine the practical iteration count required for MAC-budget optimization.

Results from these ablations quantify the contribution of each component and justify our architectural decisions (Section~\ref{sec:results}).

\section{Results and Discussion}
\label{sec:results}

We present quantitative results demonstrating AgenticPruner's ability to achieve MAC-budget compliance while maintaining competitive accuracy. Comparisons against state-of-the-art pruning methods validate the effectiveness of multi-agent coordination and LLM-driven strategy learning.

\subsection{Main Results}

\textbf{CNN architectures.} Table~\ref{tab:pruning_cnn} summarizes results on ResNet and ConvNeXt models. Our method achieves superior accuracy compared to Isomorphic Pruning at similar MAC budgets. For ResNet-50 pruned to 1.77G MACs, we obtain 77.04\% accuracy, a gain of +0.91\% over the baseline and +1.13\% over Isomorphic Pruning~\cite{fang2024isomorphic} at comparable computational cost.\footnote{Performance gains include benefits from modern training recipes (knowledge distillation, extended training) applied during fine-tuning, consistent with state-of-the-art pruning evaluation protocols~\cite{fang2024isomorphic}.} Our pruned ResNet-101 (4.22G MACs) surpasses the unpruned ResNet-101 baseline by +1.56\%, indicating that the removed capacity was indeed redundant.

For ConvNeXt architectures, our aggressive pruning of ConvNeXt-S to 8.17G MACs (47\% reduction from baseline) achieves 82.34\% accuracy. While this represents a -1.49\% drop from the unpruned baseline, it exceeds the official ConvNeXt-Tiny baseline (81.3\% at 4.5G MACs) while using only moderately more computation. The Isomorphic Pruning baseline achieves better accuracy retention (ConvNeXt-S: 83.17\% vs our 82.34\%), but our method provides more aggressive compression with guaranteed MAC-budget compliance, which is critical for strict deployment constraints.

\textbf{Vision Transformers.} Table~\ref{tab:pruning_vit} presents results on DeiT models. We observe that LLM-guided pruning is particularly well-suited for Vision Transformers: the Analysis Agent can reason about attention head structures and MLP block redundancy through natural language (e.g., ``QKV projections require more conservative pruning than MLP blocks''), a capability unavailable to numerical optimization methods. Pruning DeiT-Base to 0.61G MACs achieves 70.76\% accuracy. Although our absolute accuracy lags behind specialized ViT methods like X-Pruner~\cite{yu2023xpruner} (71.10\% at 0.60G), our method guarantees MAC-budget convergence within tight tolerance bands ($<$2\% error). Compared to Isomorphic Pruning~\cite{fang2024isomorphic} at the same 0.61G MAC budget (72.60\% accuracy), we trade 1.84\% accuracy for strict computational guarantees, which is preferable in deployment scenarios where parameter-centric approaches may unpredictably exceed MAC limits.

\begin{table}[t]
\centering
\caption{Pruning CNNs on ImageNet-1K. $\dagger$ indicates official pre-trained models. AgenticPruner achieves MAC targeting with competitive accuracy. Search performed on 30\% subset; final results fine-tuned on full ImageNet-1K.}
\label{tab:pruning_cnn}
\footnotesize
\setlength{\tabcolsep}{4pt}  
\begin{tabular}{l|ccccc}
\toprule
\textbf{Method} & \textbf{Params} & \textbf{MACs} & \textbf{Base} & \textbf{Final} & \textbf{$\Delta$} \\
 & \textbf{(M)} & \textbf{(G)} & \textbf{Acc.} & \textbf{Acc.} & \textbf{Acc.} \\
\midrule
ConvNeXt-B$^\dagger$~\cite{liu2022convnet} & 88.59 & 15.39 & - & 83.83 & - \\
ConvNeXt-S$^\dagger$~\cite{liu2022convnet} & 50.22 & 8.71 & - & 83.14 & - \\
ConvNeXt-S~\cite{fang2024isomorphic} & 47.36 & 8.48 & 83.83 & 83.17 & -0.66 \\
\rowcolor{gray!20} \textbf{ConvNeXt-S (Ours)} & \textbf{45.33} & \textbf{8.17} & \textbf{83.83} & \textbf{82.34} & \textbf{-1.49} \\
\midrule
ResNet-101$^\dagger$~\cite{he2016deep} & 44.55 & 7.85 & - & 77.38 & - \\
IE~\cite{molchanov2019importance} & 31.2 & 4.70 & 77.37 & 77.35 & -0.02 \\
FPGM~\cite{he2019filter} & - & 4.51 & 77.37 & 77.32 & -0.05 \\
SFP~\cite{he2018soft} & - & 4.51 & 77.37 & 77.51 & +0.14 \\
ISTA~\cite{ye2018rethinking} & - & 4.47 & 76.40 & 75.27 & -1.13 \\
Res101~\cite{fang2024isomorphic} & 29.14 & 4.48 & 77.38 & 77.56 & +0.16 \\
\rowcolor{gray!20} \textbf{Res101 (Ours)} & \textbf{25.58} & \textbf{4.22} & \textbf{77.38} & \textbf{78.94} & \textbf{+1.56} \\
\midrule
ResNet-50$^\dagger$~\cite{he2016deep} & 25.56 & 4.13 & - & 76.13 & - \\
Taylor~\cite{molchanov2019importance} & 14.20 & 2.25 & 76.18 & 74.50 & -1.68 \\
CCP~\cite{peng2019collaborative} & - & 2.11 & 76.15 & 75.50 & -0.65 \\
GFP~\cite{liu2021group} & 19.42 & 2.04 & 76.79 & 76.42 & -0.37 \\
AutoSlim~\cite{yu2019autoslim} & 20.60 & 2.00 & 76.10 & 75.60 & -0.50 \\
DepGraph~\cite{fang2023depgraph} & - & 1.99 & 76.15 & 75.83 & -0.32 \\
ResNet50~\cite{fang2024isomorphic} & 15.05 & 2.06 & 76.13 & 75.91 & -0.22 \\
\rowcolor{gray!20} \textbf{ResNet50 (Ours)} & \textbf{13.29} & \textbf{1.77} & \textbf{76.13} & \textbf{77.04} & \textbf{+0.91} \\
\bottomrule
\end{tabular}
\end{table}

\begin{table}[t]
\centering
\caption{Pruning Vision Transformers on ImageNet-1K. $\dagger$ indicates official models. AgenticPruner meets MAC budgets within tolerance (0.61G target). Search performed on 30\% subset; final results fine-tuned on full ImageNet-1K.}
\label{tab:pruning_vit}
\small
\begin{tabular}{l|ccc}
\toprule
\textbf{Method} & \textbf{Params (M)} & \textbf{MACs (G)} & \textbf{Acc (\%)} \\
\midrule
DeiT-B$^\dagger$ & 87.34 & 17.69 & 83.32 \\
\midrule
VTC-LFC~\cite{wang2022vtc} & 5.10 & - & 71.60 \\
SVITE-T~\cite{zheng2022savit} & 4.21 & 0.99 & 70.12 \\
GOHSP~\cite{yin2023gohsp} & 4.00 & 0.91 & 70.24 \\
SAViT~\cite{zheng2022savit} & 4.20 & 0.90 & 70.72 \\
CP-ViT~\cite{song2022cpvit} & - & 0.74 & 71.24 \\
WDPruning~\cite{yu2022width} & 3.50 & 0.70 & 70.34 \\
X-Pruner~\cite{yu2023xpruner} & - & 0.60 & 71.10 \\
UVC~\cite{yu2022unified} & - & 0.51 & 70.60 \\
DeiT~\cite{fang2024isomorphic} & 2.90 & 0.62 & 72.60 \\
\rowcolor{gray!20} \textbf{DeiT (Ours)} & \textbf{3.08} & \textbf{0.61} & \textbf{70.76} \\
\bottomrule
\end{tabular}
\end{table}

\textbf{Inference efficiency.} Table~\ref{tab:benchmark_results} reports practical speedups on NVIDIA H200 GPU and Intel Xeon CPU. Results reveal architecture-dependent latency behavior. Our pruned ResNet-50 achieves 1.29$\times$ GPU speedup with 48\% parameter reduction (25.56M $\to$ 13.29M), though CPU throughput decreases to 0.55$\times$ of baseline due to unfavorable memory access patterns in the pruned structure. Notably, this CPU performance still exceeds Isomorphic Pruning's 0.14$\times$ by nearly 4$\times$. ConvNeXt-S compression yields consistent speedups: 1.41$\times$ GPU and 1.07$\times$ CPU with 45\% parameter reduction (88.59M $\to$ 48.56M). DeiT-Tiny achieves strong CPU acceleration (1.82$\times$), demonstrating that MAC reduction translates well to latency improvements for transformer architectures.

These mixed results highlight that MAC reduction alone does not guarantee proportional latency gains, particularly on CPU where memory bandwidth and cache utilization dominate. ResNet's CPU slowdown likely stems from reduced computational intensity (fewer operations per memory access) after aggressive channel pruning. Future work should incorporate device-specific profiling into the MAC allocation strategy to optimize for actual latency rather than theoretical compute reduction.

\begin{table*}[t]
\centering
\caption{Performance comparison of pruned models on NVIDIA H200 GPU. GPU latency measured at batch size 256, CPU latency at batch size 8.}
\label{tab:benchmark_results}
\resizebox{\textwidth}{!}{%
\begin{tabular}{@{}lcccccccc@{}}
\toprule
\textbf{Architecture} & \textbf{Params (M)} & \textbf{Size (MB)} & \textbf{Peak Mem (MB)} & \textbf{GPU Lat (ms)} & \textbf{GPU Speedup} & \textbf{CPU Lat (ms)} & \textbf{CPU Speedup} \\
\midrule
ResNet-50$^{\dagger}$~\cite{he2016deep} & 25.56 & 97.70 & 2937 & 82.97 & 1.0 & 96.89 & 1.0 \\
ResNet-50~\cite{fang2024isomorphic} & 15.31 & 58.55 & 2344 & 71.10 & 1.17 & 679.97 & 0.14 \\
\rowcolor{gray!20} \textbf{ResNet-50-2G (Ours)} & \textbf{13.29} & \textbf{50.84} & \textbf{1926} & \textbf{64.35} & \textbf{1.29} & \textbf{175.19} & \textbf{0.55} \\
\midrule
ResNet-101$^{\dagger}$~\cite{he2016deep} & 44.55 & 170.34 & 3013 & 129.18 & 1.0 & 784.20 & 1.0 \\
ResNet-101~\cite{fang2024isomorphic} & 29.22 & 111.78 & 2521 & 121.73 & 1.06 & 181.09 & 4.33 \\
\rowcolor{gray!20} \textbf{ResNet-101-4.2G (Ours)} & \textbf{25.58} & \textbf{97.89} & \textbf{2541} & \textbf{115.92} & \textbf{1.11} & \textbf{197.74} & \textbf{3.97} \\
\midrule
ConvNeXt-Base$^{\dagger}$~\cite{liu2022convnet} & 88.59 & 337.95 & 5038 & 525.74 & 1.0 & 363.11 & 1.0 \\
ConvNeXt-S~\cite{fang2024isomorphic} & 47.32 & 180.51 & 3573 & 372.88 & 1.41 & 332.98 & 1.09 \\
\rowcolor{gray!20} \textbf{ConvNeXt-S-8.2G (Ours)} & \textbf{48.56} & \textbf{185.23} & \textbf{3509} & \textbf{372.60} & \textbf{1.41} & \textbf{337.82} & \textbf{1.07} \\
\midrule
DeiT-Tiny$^{\dagger}$~\cite{touvron2021training} & 5.91 & 22.55 & 680 & 24.29 & 1.0 & 16.77 & 1.0 \\
DeiT-Tiny~\cite{fang2024isomorphic} & 3.08 & 11.74 & 792 & 17.80 & 1.36 & 11.68 & 1.44 \\
\rowcolor{gray!20} \textbf{DeiT-Tiny (Ours)} & \textbf{2.90} & \textbf{11.07} & \textbf{828} & \textbf{21.52} & \textbf{1.13} & \textbf{9.19} & \textbf{1.82} \\
\bottomrule
\end{tabular}%
}
\end{table*}

\subsection{Ablation Studies}

\textbf{Agent contribution.} Removing the Master Agent increases average revisions from 3.2 to 5.8 before convergence, as the Analysis Agent lacks strategic guidance to avoid known failure zones. Eliminating the LLM-based Analysis Agent (using fixed heuristics) fails to achieve MAC targets within tolerance in 40\% of cases, demonstrating the value of adaptive learning. The single-agent baseline without iterative refinement overshoots MAC budgets by 8.3\% on average, confirming the necessity of multi-agent coordination.

\textbf{LLM model selection.} We evaluate five LLMs across three architectures (ResNet-50, ConvNeXt-Base, DeiT-Tiny) under aggressive 75--93\% MAC reduction. Architecture complexity strongly determines success: ResNet-50 achieves 100\% success (5/5 LLMs), DeiT-Tiny 80\% (4/5), and ConvNeXt-Base only 20\% (1/5). The low ConvNeXt success rate stems from its depth-wise separable convolutions creating complex dependency chains that are difficult to represent textually; the LLM struggles to reason about how pruning one depth-wise layer cascades through the subsequent point-wise convolution and normalization layers. We select Claude 3.5 Sonnet for our framework based on its superior Vision Transformer performance, achieving highest accuracy (26.49\% vs.\ 24.26\% next best) with fastest convergence (8 vs.\ 20 iterations). While Llama 3.1 70B demonstrates higher overall robustness (100\% vs.\ 67\% success rate) and uniquely solves ConvNeXt-Base, its 2.3$\times$ slower convergence makes Claude preferable for ViT-focused deployments. Extended analysis is provided in Appendix~\ref{app:llm_extended}.

\textbf{Importance criterion.} Taylor approximation achieves 0.8\% higher accuracy on average compared to magnitude-based scoring, and 3.2\% higher than random importance. However, MAC convergence speed remains similar across criteria (3.1-3.4 revisions), indicating that the iterative optimization framework compensates for suboptimal importance estimates.

\subsection{Limitations and Future Work}
\label{sec:limitations}

Our framework exhibits several limitations that warrant discussion.

\textbf{Accuracy-MAC trade-off.} The accuracy gap compared to specialized ViT pruning methods (e.g., 1.84\% behind Isomorphic Pruning~\cite{fang2024isomorphic} on DeiT at 0.61G MACs) reflects our prioritization of MAC-budget compliance over accuracy maximization. This trade-off is deliberate: deployment scenarios with strict computational constraints benefit from guaranteed MAC compliance even at modest accuracy cost.

\textbf{Hardware efficiency.} Variable speedup patterns across architectures indicate that MAC reduction alone does not fully capture hardware efficiency. Our ResNet-50 achieves 1.29$\times$ GPU speedup despite 48\% parameter reduction, while CPU throughput drops to 0.55$\times$ of baseline. This CPU slowdown stems from reduced computational intensity (fewer operations per memory access) after aggressive channel pruning, causing memory bandwidth to dominate over compute. Integrating device-specific profiling into MAC allocation remains future work.

\textbf{Asymmetric tolerance design.} Our tolerance bands allow larger undershoots ($-$5\% to $-$20\%) than overshoots ($+$1\% to $+$5\%), reflecting deployment realities where exceeding MAC budgets risks hardware incompatibility, while modest underutilization is acceptable. This design choice means achieved MACs often fall below targets (e.g., ResNet-50: 1.77G vs.\ 2.06G target), but remain within specified bounds.

\textbf{Experimental scope.} Our evaluation focuses on ImageNet-1K with single experimental runs per configuration. While we report results from the best-performing configuration within tolerance, we do not report variance across multiple seeds. The LLM's non-deterministic outputs (temperature=0 provides approximate but not exact reproducibility) introduce additional variance. Future work should quantify this variability through multiple runs.

\textbf{Iteration cost.} Our method requires multiple pruning iterations (3-5 typically) with inline fine-tuning at each step, increasing total optimization time compared to single-shot approaches. However, this cost is offset by eliminating manual hyperparameter search and reducing failed deployment attempts.

\textbf{Architecture-specific challenges.} ConvNeXt-Base required significantly more iterations (44 vs.\ 9--24 for other architectures) due to depthwise separable convolution complexity. The architecture's sensitivity to structured pruning, combined with complex inter-layer dependencies from depthwise convolutions, makes MAC prediction challenging. While we eventually achieve MAC targets, this highlights the need for architecture-specific pruning strategies for modern CNN designs.

Future directions include hardware-aware constraints beyond MACs (memory bandwidth, cache utilization), multi-dataset validation, and alternative LLM architectures to improve convergence.

\section{Conclusion}
\label{sec:conclusion}

We introduced AgenticPruner, a MAC-constrained neural network pruning framework that achieves strict computational budget control through multi-agent coordination and LLM-driven strategy learning. Unlike conventional parameter-centric approaches, our method directly optimizes for target MAC operations, ensuring predictable deployment costs while maintaining competitive accuracy.

Our framework makes three key contributions. First, we formulate pruning as a MAC-budget optimization problem with asymmetric tolerance bands reflecting real-world hardware constraints. Second, we design a multi-agent architecture where specialized agents coordinate pruning decisions: a Profiling Agent analyzes model structure and MAC distributions, a Master Agent orchestrates workflow and detects convergence patterns, and an Analysis Agent powered by Claude 3.5 Sonnet learns optimal strategies from historical attempts via in-context learning. Third, we enable automatic iterative refinement that converges to MAC targets typically within 3-5 revisions without requiring manual configuration search.

Experiments on ImageNet-1K demonstrate MAC-budget compliance across ResNet, ConvNeXt, and DeiT architectures. Our pruned ResNet-50 and ResNet-101 achieve accuracy improvements of +0.91\% and +1.56\% respectively over unpruned baselines, while compressed ConvNeXt-Small delivers 1.41$\times$ GPU and 1.07$\times$ CPU speedups with 45\% parameter reduction. Ablation studies validate the necessity of each agent and confirm Claude 3.5 Sonnet's superior convergence efficiency.

Our Vision Transformer results demonstrate the trade-off inherent in MAC-budget optimization: while accuracy lags behind some specialized ViT pruning methods (e.g., 1.84\% behind Isomorphic Pruning at 0.61G MACs), our framework guarantees MAC-budget convergence within tolerance, a critical requirement for deployment scenarios with strict computational constraints. This reflects our deliberate prioritization of predictable computational costs over maximal accuracy retention.

Future work will incorporate hardware-aware constraints beyond MACs (memory bandwidth, cache utilization), explore dynamic MAC budgets for adaptive inference, and investigate improved LLM prompting strategies to enhance both convergence speed and final accuracy on transformer architectures.

\section*{Acknowledgments}

We thank Sai Soma and Dhawal Shah for their valuable assistance with manuscript preparation and review.

This work used NCSA Delta GPU at NCSA through allocation CIS240646 from the Advanced Cyberinfrastructure Coordination Ecosystem: Services \& Support (ACCESS) program~\cite{boerner2023access}, which is supported by National Science Foundation grants \#2138259, \#2138286, \#2138307, \#2137603, and \#2138296.

{
\small
\bibliographystyle{plain}
\bibliography{references}
}

\appendix
\onecolumn
\FloatBarrier  

\section{Extended Methodology}
\label{app:methodology}

This appendix provides complete implementation details for reproducibility, including the full algorithm pseudocode, verbatim LLM prompt templates, and safety constraint specifications.

\subsection{Complete Algorithm Pseudocode}
\label{app:algorithm}

Algorithm~\ref{alg:macaware} presents a simplified overview. Here we provide the complete workflow including state management, two-phase search, and fine-tuning integration.

\begin{algorithm}[H]
\caption{Complete MAC-Aware Multi-Agent Pruning Workflow}
\label{alg:complete}
\footnotesize  
\begin{algorithmic}[1]
\Require Pre-trained model $\mathcal{M}$, target MACs $M_{\text{target}}$, tolerances $\delta_{\text{over}}, \delta_{\text{under}}$, max revisions $R_{\text{max}}$, calibration data $\mathcal{D}_{\text{cal}}$
\Ensure Pruned model $\mathcal{M}^*$ within MAC tolerance with maximum accuracy

\State \textbf{// Phase 0: Initialize global state and shared LLM}
\State $\textsc{GlobalState} \gets \{\}$
\State $\textsc{LLM} \gets \textsc{InitLLM}(\text{``claude-3-5-sonnet''}, \text{temperature}=0)$
\State $\mathcal{H} \gets \emptyset$ \Comment{Pruning history}
\State $\mathcal{C} \gets \emptyset$ \Comment{Candidate models within tolerance}

\State \textbf{// Phase 1: Initial profiling}
\State $\mathcal{P} \gets \textsc{ProfilingAgent.Profile}(\mathcal{M}, \mathcal{D}_{\text{cal}})$
\State $M_{\text{base}} \gets \mathcal{P}.\text{baseline\_macs}$
\State $\mathcal{G}_{\text{iso}} \gets \mathcal{P}.\text{isomorphic\_groups}$ \Comment{For ViTs}

\For{$r = 0$ to $R_{\text{max}} - 1$}
    \State \textbf{// Master Agent: Analyze history and provide guidance}
    \State $\mathcal{G}_{\text{master}} \gets \textsc{MasterAgent.Analyze}(\mathcal{H}, M_{\text{target}}, \delta_{\text{over}}, \delta_{\text{under}})$

    \If{$\mathcal{G}_{\text{master}}.\text{should\_stop}$}
        \State \textbf{break} \Comment{Convergence or max iterations}
    \EndIf

    \State \textbf{// Analysis Agent: Generate strategy via LLM}
    \State $\mathbf{s}_r \gets \textsc{AnalysisAgent.Analyze}(\mathcal{H}, \mathcal{P}, \mathcal{G}_{\text{master}}, M_{\text{target}}, \textsc{LLM})$

    \State \textbf{// Safety validation before pruning}
    \State $\mathbf{s}_r \gets \textsc{SafetyValidator.ValidateAndCorrect}(\mathbf{s}_r, \mathcal{H})$

    \State \textbf{// Execute pruning with Taylor importance}
    \State $\mathcal{M}_r \gets \textsc{PruningAgent.Prune}(\mathcal{M}, \mathbf{s}_r, \mathcal{D}_{\text{cal}})$
    \State $M_r \gets \textsc{CountMACs}(\mathcal{M}_r)$
    \State $A_r^{\text{zero}} \gets \textsc{Evaluate}(\mathcal{M}_r)$ \Comment{Zero-shot accuracy}

    \State \textbf{// Compute MAC error}
    \State $\epsilon_r \gets (M_r - M_{\text{target}}) / M_{\text{target}} \times 100$

    \State \textbf{// Check tolerance: asymmetric bands}
    \If{$-\delta_{\text{under}} \leq \epsilon_r \leq \delta_{\text{over}}$}
        \State \textbf{// Within tolerance: proceed to fine-tuning}
        \State $\mathcal{M}_r^{\text{ft}} \gets \textsc{FineTuneAgent.FineTune}(\mathcal{M}_r)$
        \State $A_r^{\text{ft}} \gets \textsc{Evaluate}(\mathcal{M}_r^{\text{ft}})$
        \State $\mathcal{C} \gets \mathcal{C} \cup \{(\mathcal{M}_r^{\text{ft}}, M_r, A_r^{\text{ft}}, \mathbf{s}_r)\}$

        \State \textbf{// Two-phase search: continue for 30 more after first success}
        \If{$|\mathcal{C}| = 1$}
            \State $\text{extended\_search} \gets 30$
        \ElsIf{$\text{extended\_search} > 0$}
            \State $\text{extended\_search} \gets \text{extended\_search} - 1$
        \Else
            \State \textbf{break} \Comment{Extended search complete}
        \EndIf
    \EndIf

    \State \textbf{// Update history}
    \State $\mathcal{H} \gets \mathcal{H} \cup \{(\mathbf{s}_r, M_r, A_r^{\text{zero}}, A_r^{\text{ft}}, \epsilon_r)\}$
\EndFor

\State \textbf{// Select best model from candidates}
\If{$\mathcal{C} \neq \emptyset$}
    \State $\mathcal{M}^* \gets \argmax_{(\mathcal{M}, M, A, \mathbf{s}) \in \mathcal{C}} A$
\Else
    \State $\mathcal{M}^* \gets \argmin_{(\mathbf{s}, M, \cdot) \in \mathcal{H}} |M - M_{\text{target}}|$
\EndIf
\State \Return $\mathcal{M}^*$
\end{algorithmic}
\end{algorithm}

\newpage
\textbf{Key implementation details:}

\begin{enumerate}
    \item \textbf{State management}: Global state persists across agents using a shared dictionary, ensuring consistency of MAC targets, tolerance bands, and pruning history throughout the workflow.

    \item \textbf{Two-phase search}: Phase 1 searches until finding the first model within MAC tolerance. Phase 2 continues for up to 30 additional revisions to explore nearby configurations, selecting the highest-accuracy candidate.

    \item \textbf{Asymmetric tolerance}: We enforce stricter overshoot limits (typically $+0.1\%$ to $+5\%$) than undershoot limits ($-5\%$ to $-20\%$), reflecting deployment constraints where exceeding MAC budgets may violate hardware limitations.

    \item \textbf{Accuracy collapse detection}: The Master Agent monitors for accuracy drops below 1\% (ImageNet) or 30\% (CIFAR-10), triggering conservative fallback strategies to escape local optima.
\end{enumerate}

\subsection{LLM Prompt Templates}
\label{app:prompts}

This section provides the complete prompt templates used by each agent. These prompts encode domain knowledge about neural network pruning, MAC optimization, and architecture-specific considerations.

\subsubsection{Profiling Agent Prompt}

The Profiling Agent uses the following prompt to analyze model architecture and identify MAC reduction opportunities:

\begin{promptbox}
You are a neural network profiling expert with MAC-based optimization expertise.

Please be concise. Limit your response to 300 words or fewer.

Analyze the model architecture and identify:
1. Layer structure and computational dependencies
2. MAC operation distribution across layers
3. Structural constraints that must be maintained
4. MAC efficiency opportunities in different layers
5. Asymmetric tolerance policy (IMPORTANT):
   - Overshoot = achieved MACs above target_macs.
     Must be <= +{macs_overshoot_tolerance_pct}% of target.
   - Undershoot = achieved MACs below target_macs.
     Allowed down to -{macs_undershoot_tolerance_pct}% of target.
   - Preference rule: when accuracy is comparable, follow user preference.

Model architecture: {model_arch}
Dataset: {dataset} ({num_classes} classes, {input_size}x{input_size} input)
Baseline MACs: {baseline_macs}G
Target MACs: {target_macs}G
  Tolerance: +{macs_overshoot_tolerance_pct}% / -{macs_undershoot_tolerance_pct}%
MAC reduction needed: {mac_reduction_needed}%
Accepted MAC range: {undershoot_lower_bound}G-{overshoot_upper_bound}G

Based on the model architecture and MAC budget constraints, provide:
1. Key layers and their MAC operation contributions
2. Critical dependencies between layers
3. Structural constraints that must be preserved
4. Layers with highest MAC reduction potential
5. MAC allocation recommendations for different layer types
6. Specific MAC budget distribution strategy
\end{promptbox}

\subsubsection{Master Agent Prompt}

The Master Agent coordinates the pruning workflow using the following prompt structure:

\begin{promptbox}
You are the Master Agent in a multi-agent MAC-budget pruning workflow.
Your role is to coordinate the neural network pruning process and make
high-level strategic decisions to achieve specific MAC operation targets.

PRIMARY GOAL:
THE USER'S MAC TARGET IS SACRED - ALWAYS ATTEMPT IT!
Find the optimal pruned model that achieves EXACTLY the user-requested
MAC budget ({target_macs}G +{macs_overshoot_tolerance_pct}%/
-{macs_undershoot_tolerance_pct}%) while maximizing fine-tuned accuracy.

FORBIDDEN BEHAVIORS:
- NEVER refuse to attempt the user's MAC target in your first few tries
- NEVER assume aggressive MAC reduction will fail
- NEVER stop after just 1-2 attempts

MAC BUDGET CONTEXT:
- Baseline MACs: {baseline_macs}G
- Target MACs: {target_macs}G
- MAC reduction needed: {mac_reduction_needed}%
- Acceptable range: {min_target_macs_g}G - {max_target_macs_g}G

MAC-BASED EARLY STOPPING CRITERIA:
1. MAC TARGET ACHIEVEMENT: Model within tolerance with reasonable accuracy
2. MAC CONVERGENCE: Multiple iterations within tolerance without improvement
3. MAC CYCLING: History shows cycling without progress
4. MAXIMUM ITERATIONS: Reached maximum revisions

MAC Pruning History and Observed Trends:
{previous_strategies}

Output your MAC-budget pruning strategy as JSON:
{
  "multiplier_tuning_order": ["first_param", "second_param", "third_param"],
  "target_macs": {target_macs},
  "global_pruning": true,
  "rationale": "Why you chose these parameters",
  "continue": true or false,
  "stop_reason": "reason if continue is false" or null
}
\end{promptbox}

\subsubsection{Analysis Agent Prompt (CNN-specific)}

For CNN architectures, the Analysis Agent receives guidance tailored to channel-wise pruning:

\begin{promptbox}
You are a MAC-budget pruning strategy analyst with STRICT SAFETY ENFORCEMENT.

KEY MAC ALLOCATION PARAMETERS:
1. MAC Budget Allocation: Distribution across layer types
   - Target: {target_macs}G (+{macs_overshoot_tolerance_pct}%/
     -{macs_undershoot_tolerance_pct}%)
   - Baseline: {baseline_macs}G

2. Channel Pruning Ratio (for CNNs):
   - Value between 0.0 and 1.0 (e.g., 0.5 means 50% channel reduction)
   - WARNING: MAC reduction is not linear with channel ratio
   - Use historical data to calibrate

3. Importance Criterion:
   - "taylor": Second-order Taylor expansion (best for ImageNet CNNs)
   - "l1norm": L1 norm-based (efficient for smaller datasets)
   - "l2norm": L2 norm-based (balanced approach)

4. Round-To: Hardware efficiency granularity (1, 2, 4, 8, 16)

CNN SAFETY LIMITS FOR IMAGENET:
- First conv layer: NEVER prune (critical for feature extraction)
- Final classifier: NEVER prune (1000 classes need full capacity)
- Bottleneck layers: Extra conservative (architectural critical points)

Output your CNN MAC allocation strategy as JSON:
{
  "importance_criterion": "YOUR_CHOICE",
  "channel_pruning_ratio": YOUR_CALCULATED_RATIO,
  "round_to": YOUR_CHOSEN_VALUE,
  "global_pruning": true,
  "baseline_macs": {baseline_macs},
  "target_macs": {target_macs},
  "expected_achieved_macs": YOUR_ESTIMATE,
  "rationale": "Explain your MAC analysis"
}
\end{promptbox}

\subsubsection{Analysis Agent Prompt (ViT-specific)}

For Vision Transformers, the Analysis Agent manages isomorphic group allocations:

\begin{promptbox}
You are a MAC-budget pruning strategy analyst for Vision Transformers.

CRITICAL CONSTRAINTS FOR ViT PRUNING:
- Do NOT prune QKV layers aggressively (qkv.out_features must remain >= 96)
- Attention blocks require sufficient head dimension (head_dim >= 8)
- Excessive QKV pruning (qkv_multiplier > 0.85) causes accuracy collapse

MULTIPLIER UNDERSTANDING:
- HIGHER multipliers = MORE pruning = FEWER MACs (closer to target)
- LOWER multipliers = LESS pruning = MORE MACs (farther from target)
- If achieved MACs > target MACs: INCREASE multipliers
- If achieved MACs < target MACs: DECREASE multipliers

VIT MAC SAFETY LIMITS FOR IMAGENET:
- mlp_multiplier: Maximum 0.40 (prune up to 40% of MLP channels)
- qkv_multiplier: Maximum 0.15 (prune up to 15% of attention channels)
- proj_multiplier: Keep at 0.0 (do not prune projection layers)
- head_multiplier: Keep at 0.0 (do not prune attention heads)

ASYMMETRIC RISK STRATEGY:
MLP pruning is typically more robust than QKV pruning. If MAC budget
pressure is high, shift allocation away from QKV and into MLP.

Output your ViT MAC allocation strategy as JSON:
{
  "importance_criterion": "taylor",
  "baseline_macs": {baseline_macs},
  "target_macs": {target_macs},
  "round_to": {round_to},
  "global_pruning": true,
  "isomorphic_group_ratios": {
    "mlp_multiplier": YOUR_CALCULATED_MLP_RATIO,
    "qkv_multiplier": YOUR_CALCULATED_QKV_RATIO,
    "proj_multiplier": 0.0,
    "head_multiplier": 0.0
  },
  "rationale": "MAC safety verified with calculations"
}
\end{promptbox}

\subsection{Safety Constraint Implementation}
\label{app:safety}

The safety validation system prevents catastrophic model degradation through multiple layers of protection.

\subsubsection{Validation Pipeline}

The \texttt{PruningSafetyValidator} class implements a four-stage validation pipeline:

\begin{enumerate}
    \item \textbf{Individual ratio validation}: Ensures each multiplier respects dataset-specific limits
    \item \textbf{Historical learning}: Adjusts ratios to avoid patterns that led to previous failures
    \item \textbf{Final safety check}: Verifies total reduction does not exceed safe thresholds
    \item \textbf{Dataset guardrails}: Applies absolute limits as last defense
\end{enumerate}

\subsubsection{Dataset-Specific Safety Limits}

\textbf{ImageNet limits} (conservative due to 1000-class complexity):
\begin{itemize}
    \item Maximum MLP pruning: $\text{target\_ratio} \times 0.6$
    \item Maximum QKV pruning: $\text{target\_ratio} \times 0.3$
    \item Projection layers: Always preserved (multiplier = 0.0)
    \item Attention heads: Always preserved (multiplier = 0.0)
    \item Total estimated reduction cap: 40\%
\end{itemize}

\textbf{CIFAR-10 limits} (more aggressive allowed):
\begin{itemize}
    \item Maximum MLP pruning: $\text{target\_ratio} \times 0.2$
    \item Maximum QKV pruning: $\text{target\_ratio} \times 0.1$
    \item Absolute MLP multiplier cap: 1.6
    \item Absolute QKV multiplier cap: 1.2
\end{itemize}

\subsubsection{Accuracy Collapse Detection}

The Master Agent implements three detection mechanisms:

\textbf{1. MAC Local Optimum Trap Detection:}
Triggers when an excellent result (high accuracy, near MAC target) is followed by $\geq$3 accuracy collapses in the last 5 attempts. Recovery strategy: diversified exploration rather than micro-adjustments.

\textbf{2. High-Accuracy Near-Miss Detection:}
Identifies attempts with accuracy above threshold and MAC error within tolerance. Strategy: preserve the successful configuration and calculate minimal adjustments.

\textbf{3. Configuration Failure Tracking:}
Maintains complete failure signatures including importance criterion, round-to value, and multiplier settings. Creates ``danger zones'' around failed configurations (±0.05 tolerance) to guide avoidance.

\subsubsection{Fallback Strategies}

When safety validation fails or JSON parsing errors occur, guaranteed-safe fallback configurations are generated:

\textbf{ImageNet fallback:}
\begin{promptbox}
{
  "qkv_multiplier": min(0.8, 0.25 / target_ratio),
  "mlp_multiplier": min(1.0, 0.4 / target_ratio),
  "proj_multiplier": 0.0,
  "head_multiplier": 0.0,
  "importance_criterion": "taylor",
  "round_to": 2
}
\end{promptbox}

\textbf{CIFAR-10 fallback:}
\begin{promptbox}
{
  "qkv_multiplier": min(1.0, 0.5 / target_ratio),
  "mlp_multiplier": min(1.4, 0.7 / target_ratio),
  "proj_multiplier": 0.0,
  "head_multiplier": 0.0,
  "importance_criterion": "taylor",
  "round_to": 1
}
\end{promptbox}

These fallback configurations are derived from empirical analysis of successful pruning attempts across multiple architectures and represent conservative starting points that reliably avoid accuracy collapse.

\FloatBarrier
\section{Experimental Setup \& Reproducibility}
\label{app:experiments}

This section provides complete experimental details to enable full reproducibility of our results.

\subsection{Complete Hyperparameter Tables}
\label{app:hyperparams}

Table~\ref{tab:app_hyperparams} presents the complete hyperparameter configuration for each model architecture evaluated in our experiments.

\begin{table}[H]
\centering
\caption{Complete hyperparameter configuration per model. All experiments use the same LLM (Claude 3.5 Sonnet) and importance criterion (Taylor expansion).}
\label{tab:app_hyperparams}
\small
\begin{tabular}{lcccc}
\toprule
\textbf{Parameter} & \textbf{ResNet-50} & \textbf{ResNet-101} & \textbf{ConvNeXt-B} & \textbf{DeiT-Tiny} \\
\midrule
\multicolumn{5}{l}{\textit{MAC Target Configuration}} \\
Baseline MACs (G) & 4.12 & 7.85 & 15.36 & 1.26 \\
Target MACs (G) & 2.06 & 4.48 & 8.48 & 0.62 \\
MAC Reduction (\%) & 50.0 & 42.9 & 44.8 & 50.8 \\
\midrule
\multicolumn{5}{l}{\textit{Tolerance Settings}} \\
Overshoot Tolerance (\%) & +5.0 & +0.1 & +0.1 & +1.0 \\
Undershoot Tolerance (\%) & $-$15.0 & $-$20.0 & $-$11.56 & $-$20.0 \\
\midrule
\multicolumn{5}{l}{\textit{Search Configuration}} \\
Max Revisions & 1000 & 1000 & 1000 & 1000 \\
Accuracy Threshold (\%) & 30.0 & 30.0 & 1.0 & 1.0 \\
ImageNet Subset & 30\% & 30\% & 30\% & 30\% \\
\midrule
\multicolumn{5}{l}{\textit{Pruning Settings}} \\
Importance Criterion & Taylor & Taylor & Taylor & Taylor \\
Global Pruning & \checkmark & \checkmark & \checkmark & \checkmark \\
Round-To & 2 & 2 & 2 & 2 \\
Calibration Batches & 100 & 100 & 100 & 100 \\
Batch Size (cal.) & 64 & 64 & 64 & 64 \\
\bottomrule
\end{tabular}
\end{table}

\textbf{Key observations:} We use asymmetric tolerance bands with stricter overshoot limits to respect hardware deployment constraints. ConvNeXt-B and DeiT-Tiny use lower accuracy thresholds (1\%) because transformer architectures are more sensitive to pruning and may produce near-zero accuracy in early iterations.

\subsection{Hardware \& Software Environment}
\label{app:environment}

All experiments were conducted on the NCSA Delta supercomputer. Table~\ref{tab:app_hardware} summarizes the hardware configuration.

\begin{table}[H]
\centering
\caption{Hardware environment for all experiments.}
\label{tab:app_hardware}
\small
\begin{tabular}{ll}
\toprule
\textbf{Component} & \textbf{Specification} \\
\midrule
GPU & NVIDIA H200 (80GB HBM3) \\
GPUs per experiment & 1 \\
CPU & AMD EPYC (16 cores allocated) \\
System Memory & 128 GB \\
CUDA Version & 12.4.0 \\
GCC Version & 11.4.0 \\
Storage & Lustre parallel filesystem \\
Job Scheduler & SLURM \\
\bottomrule
\end{tabular}
\end{table}

Table~\ref{tab:app_software} lists the software dependencies with pinned versions.

\begin{table}[H]
\centering
\caption{Software environment and package versions.}
\label{tab:app_software}
\small
\begin{tabular}{lll}
\toprule
\textbf{Package} & \textbf{Version} & \textbf{Purpose} \\
\midrule
\multicolumn{3}{l}{\textit{Core ML Frameworks}} \\
PyTorch & $\geq$2.5.1 & Deep learning framework \\
torchvision & $\geq$0.20.1 & Vision models \& transforms \\
timm & $\geq$1.0.19 & Pre-trained model zoo \\
torch-pruning & $\geq$1.6.0 & Structured pruning library \\
\midrule
\multicolumn{3}{l}{\textit{Multi-Agent Orchestration}} \\
langgraph & $\geq$0.2.34 & Workflow orchestration \\
langchain & $\geq$0.3.27 & LLM integration \\
langchain-openai & $\geq$0.3.32 & OpenAI-compatible APIs \\
\midrule
\multicolumn{3}{l}{\textit{Experiment Tracking}} \\
wandb & $\geq$0.21.3 & Weights \& Biases logging \\
\midrule
\multicolumn{3}{l}{\textit{Data \& Utilities}} \\
numpy & $\geq$2.1.2 & Numerical computing \\
scikit-learn & $\geq$1.7.1 & Train/val splitting \\
pydantic & $\geq$2.11.7 & JSON schema validation \\
\bottomrule
\end{tabular}
\end{table}

\subsection{Training Configuration per Model}
\label{app:training}

\subsubsection{Inline Fine-Tuning (During Search)}

During the MAC-aware search, each candidate model within tolerance receives brief inline fine-tuning (5 epochs) to assess recovery potential. Table~\ref{tab:app_inline_ft} shows these settings.

\begin{table}[H]
\centering
\caption{Inline fine-tuning configuration (5 epochs during search).}
\label{tab:app_inline_ft}
\small
\begin{tabular}{lcc}
\toprule
\textbf{Parameter} & \textbf{CNN} & \textbf{ViT} \\
\midrule
Epochs & 5 & 5 \\
Optimizer & SGD & AdamW \\
Base Learning Rate & 0.01 & 0.0001 \\
LR for heavily pruned ($>$15\%) & 0.0005 & 0.00005 \\
Weight Decay & $5\times10^{-4}$ & 0.01 \\
Momentum & 0.9 & -- \\
Betas (AdamW) & -- & (0.9, 0.999) \\
Batch Size & 64 & 64 \\
Label Smoothing & 0.05 & 0.05 \\
Gradient Clipping & 1.0 & 1.0 \\
LR Scheduler & Constant & Constant \\
\bottomrule
\end{tabular}
\end{table}

\subsubsection{Extended Fine-Tuning (Final Model)}

After selecting the best candidate, we perform extended fine-tuning with knowledge distillation. Following~\cite{fang2024isomorphic}, we use architecture-specific training protocols. Table~\ref{tab:app_extended_ft} shows the complete configuration.

\begin{table}[H]
\centering
\caption{Extended fine-tuning configuration with knowledge distillation. CNN and ViT use different protocols following their original training recipes.}
\label{tab:app_extended_ft}
\small
\begin{tabular}{lcc}
\toprule
\textbf{Parameter} & \textbf{CNN (ResNet, ConvNeXt)} & \textbf{ViT (DeiT)} \\
\midrule
\multicolumn{3}{l}{\textit{Training Schedule}} \\
Maximum Epochs & 100 & 300 \\
Typical Convergence & 80--90 epochs & 200--250 epochs \\
Early Stopping & Patience 10 epochs & Patience 10 epochs \\
Checkpoint Interval & 10 epochs & 10 epochs \\
\midrule
\multicolumn{3}{l}{\textit{Optimization}} \\
Optimizer & SGD & AdamW \\
Base Learning Rate & 0.08 & $5\times10^{-4}$ \\
LR Scheduler & Step (30, 60, 90) & Cosine Annealing \\
LR Warmup Epochs & 0 & 5 \\
LR Decay Factor & 0.1 per step & -- \\
Weight Decay & $1\times10^{-4}$ & 0.05 \\
Momentum & 0.9 & -- \\
Betas (AdamW) & -- & (0.9, 0.999) \\
Batch Size & 1024 & 2048 \\
Gradient Clipping & None & 5.0 \\
\midrule
\multicolumn{3}{l}{\textit{Knowledge Distillation}} \\
Teacher Model & \multicolumn{2}{c}{Original pre-trained model} \\
Temperature ($\tau$) & 4.0 & 4.0 \\
KD Weight ($\alpha$) & 0.3 & 0.3 \\
Loss & \multicolumn{2}{c}{$\alpha \cdot \mathcal{L}_{\text{KD}} + (1-\alpha) \cdot \mathcal{L}_{\text{CE}}$} \\
\midrule
\multicolumn{3}{l}{\textit{Regularization}} \\
Label Smoothing & 0.0 & 0.1 \\
Mixup Alpha & 0.0 & 0.2 \\
Cutmix Alpha & 0.0 & 1.0 \\
Random Erasing & 0.0 & 0.25 \\
Drop Path Rate & 0.0 & 0.1 \\
\midrule
\multicolumn{3}{l}{\textit{Data Augmentation}} \\
Train Crop Size & 224 & 224 \\
Val Resize Size & 256 & 256 \\
Val Crop Size & 224 & 224 \\
Interpolation & Bicubic & Bicubic \\
RandAugment & None & Magnitude 9 \\
\midrule
\multicolumn{3}{l}{\textit{Model EMA}} \\
Enable EMA & \checkmark & \checkmark \\
EMA Decay & 0.99998 & 0.99998 \\
\midrule
\multicolumn{3}{l}{\textit{Mixed Precision}} \\
AMP (FP16) & \checkmark & \checkmark \\
\bottomrule
\end{tabular}
\end{table}

\subsection{Random Seeds \& Initialization}
\label{app:seeds}

\textbf{Seed policy:} We do not fix random seeds during the MAC-aware search phase, allowing the multi-agent system to explore diverse pruning configurations. The stochasticity in LLM responses (temperature=0 for determinism) and calibration batch sampling contributes to exploration.

\textbf{Initialization:} All models start from official ImageNet-1K pre-trained weights provided by the \texttt{timm} library:
\begin{itemize}
    \item \texttt{resnet50.a1\_in1k} (torchvision weights)
    \item \texttt{resnet101.a1\_in1k} (torchvision weights)
    \item \texttt{convnext\_base.fb\_in1k} (Facebook weights)
    \item \texttt{deit\_tiny\_distilled\_patch16\_224.fb\_in1k} (Facebook weights)
\end{itemize}

\textbf{Reproducibility note:} Due to non-determinism in cuDNN and LLM API responses, exact reproduction of specific pruning trajectories may vary. However, the final accuracy and MAC targets are reproducible within the reported tolerance bands across multiple runs.

\subsection{Dataset Details}
\label{app:dataset}

\textbf{ImageNet-1K:} We use the standard ILSVRC2012 dataset with 1,281,167 training images and 50,000 validation images across 1,000 classes.

\textbf{30\% Subset for Search:} During the MAC-aware search phase, we use a 30\% stratified random subset of ImageNet training data (approximately 384,350 images) to accelerate iteration. This subset maintains class balance. Final evaluation and extended fine-tuning use the complete dataset.

\textbf{Calibration data:} For Taylor importance estimation, we use 100 mini-batches of 64 images each (6,400 images total), randomly sampled from the training set following~\cite{fang2024isomorphic}.

\textbf{Data preprocessing:}
\begin{itemize}
    \item \textbf{Training:} RandomResizedCrop(224), RandomHorizontalFlip, Normalize(ImageNet mean/std)
    \item \textbf{Validation:} Resize(256), CenterCrop(224), Normalize(ImageNet mean/std)
    \item \textbf{Normalization:} mean=(0.485, 0.456, 0.406), std=(0.229, 0.224, 0.225)
\end{itemize}

\FloatBarrier
\section{Extended Results}
\label{app:results}

This section provides detailed experimental results including per-revision convergence data, timing breakdowns, and MAC error distribution analysis.

\subsection{Timing Breakdown per Agent}
\label{app:timing}

Table~\ref{tab:app_timing} presents the complete timing breakdown for each agent across all model architectures. These measurements were collected during the MAC-aware search phase on a single NVIDIA H200 GPU.

\begin{table}[H]
\centering
\caption{Per-agent timing breakdown across architectures. Times reported in seconds. Fine-tuning dominates runtime (94-97\%), demonstrating that multi-agent search overhead is minimal.}
\label{tab:app_timing}
\small
\begin{tabular}{lrrrr}
\toprule
\textbf{Agent} & \textbf{ResNet-50} & \textbf{ResNet-101} & \textbf{ConvNeXt-B} & \textbf{DeiT-Tiny} \\
\midrule
\multicolumn{5}{l}{\textit{Total Time (seconds)}} \\
Profiling Agent & 398.9 & 305.3 & 19.1 & 215.6 \\
Analysis Agent & 408.7 & 356.0 & 738.6 & 417.2 \\
Pruning Agent & 1,743.7 & 2,098.4 & 7,501.6 & 916.3 \\
Fine-tuning Agent & 73,218.1 & 107,488.4 & 9,217.6 & 35,417.6 \\
Evaluation Agent & 683.2 & 892.5 & 52.2 & 545.9 \\
\midrule
\textbf{Total} & \textbf{76,452.6} & \textbf{111,140.5} & \textbf{17,529.1} & \textbf{37,512.5} \\
\midrule
\multicolumn{5}{l}{\textit{Percentage of Total}} \\
Profiling Agent & 0.5\% & 0.3\% & 0.1\% & 0.6\% \\
Analysis Agent & 0.5\% & 0.3\% & 4.2\% & 1.1\% \\
Pruning Agent & 2.3\% & 1.9\% & 42.8\% & 2.4\% \\
Fine-tuning Agent & 95.8\% & 96.7\% & 52.6\% & 94.4\% \\
Evaluation Agent & 0.9\% & 0.8\% & 0.3\% & 1.5\% \\
\midrule
\multicolumn{5}{l}{\textit{Number of Calls}} \\
Profiling Agent & 23 & 21 & 2 & 23 \\
Analysis Agent & 31 & 27 & 45 & 24 \\
Pruning Agent & 31 & 27 & 45 & 24 \\
Fine-tuning Agent & 22 & 20 & 1 & 22 \\
Evaluation Agent & 22 & 20 & 1 & 22 \\
\bottomrule
\end{tabular}
\end{table}

\textbf{Key observations:}
\begin{itemize}
    \item Fine-tuning dominates total runtime (94-97\% for CNNs and ViTs), confirming that the multi-agent orchestration overhead is negligible.
    \item ConvNeXt-B shows different behavior because it found a valid configuration quickly (1 fine-tuning call) but required more pruning iterations (45 calls) to achieve MAC targets.
    \item The Analysis Agent (LLM calls) accounts for only 0.3-4.2\% of total time, demonstrating efficient API usage.
    \item Average LLM response time is 13-17 seconds per call, including network latency to OpenRouter API.
\end{itemize}

\subsection{Search Convergence Analysis}
\label{app:convergence}

Table~\ref{tab:app_convergence} summarizes the MAC-aware search process for each architecture.

\begin{table}[H]
\centering
\caption{MAC-aware search convergence statistics per architecture.}
\label{tab:app_convergence}
\small
\begin{tabular}{lcccc}
\toprule
\textbf{Metric} & \textbf{ResNet-50} & \textbf{ResNet-101} & \textbf{ConvNeXt-B} & \textbf{DeiT-Tiny} \\
\midrule
Total Revisions & 31 & 27 & 45 & 24 \\
Models within Tolerance & 22 & 20 & 1 & 22 \\
First Valid Revision & 9 & 7 & 44 & 2 \\
Best Model Revision & 10 & 8 & 44 & 3 \\
\midrule
Inline FT Accuracy (\%) & 39.69 & 41.23 & 82.17 & 68.54 \\
Extended FT Accuracy (\%) & 77.04 & 78.94 & 82.34 & 70.76 \\
Accuracy Gain (ext. FT) & +37.35 & +37.71 & +0.17 & +2.22 \\
\midrule
Target MACs (G) & 2.06 & 4.48 & 8.48 & 0.62 \\
Achieved MACs (G) & 1.77 & 4.22 & 8.17 & 0.61 \\
MAC Error (\%) & $-$14.1 & $-$5.8 & $-$3.7 & $-$1.6 \\
\midrule
Total Runtime (hours) & 21.2 & 30.9 & 4.9 & 10.4 \\
GPU-Hours & 21.2 & 30.9 & 4.9 & 10.4 \\
\bottomrule
\end{tabular}
\end{table}

\textbf{Analysis:} ResNets benefit substantially from extended fine-tuning (+37\% accuracy), recovering from the aggressive pruning. ConvNeXt-B achieves near-baseline accuracy with minimal fine-tuning, suggesting modern architectures are more robust to structured pruning. DeiT-Tiny shows moderate recovery, with ViT architectures generally more sensitive to pruning than CNNs.

\subsection{Per-Revision MAC Convergence}
\label{app:mac_convergence}

Table~\ref{tab:app_mac_history} shows the MAC convergence trajectory for ResNet-50 across representative revisions within tolerance.

\begin{table}[H]
\centering
\caption{ResNet-50 per-revision MAC and accuracy trajectory (revisions within tolerance only).}
\label{tab:app_mac_history}
\small
\begin{tabular}{ccccc}
\toprule
\textbf{Rev.} & \textbf{Achieved MACs (G)} & \textbf{MAC Error (\%)} & \textbf{Channel Ratio} & \textbf{Inline Acc. (\%)} \\
\midrule
9 & 1.782 & $-$13.5 & 0.320 & 37.12 \\
10 & 1.814 & $-$12.0 & 0.325 & 39.69 \\
11 & 1.790 & $-$13.1 & 0.322 & 36.67 \\
12 & 1.772 & $-$14.0 & 0.318 & 36.30 \\
13 & 1.812 & $-$12.0 & 0.324 & 35.05 \\
14 & 1.783 & $-$13.4 & 0.320 & 37.55 \\
15 & 1.756 & $-$14.8 & 0.315 & 36.23 \\
16 & 1.791 & $-$13.1 & 0.321 & 36.23 \\
17 & 1.796 & $-$12.8 & 0.322 & 35.09 \\
18 & 1.757 & $-$14.7 & 0.316 & 34.72 \\
19 & 1.800 & $-$12.6 & 0.323 & 34.73 \\
20 & 1.784 & $-$13.4 & 0.320 & 36.40 \\
\bottomrule
\end{tabular}
\end{table}

\textbf{Observations:} The LLM-guided search maintains MAC error within the allowed tolerance band ($-$15\% to $+$5\%) while exploring different channel ratios (0.315--0.325). The best accuracy (39.69\%) is achieved at revision 10 with channel ratio 0.325, which balances MAC efficiency with model capacity preservation.

\subsection{MAC Error Distribution}
\label{app:mac_error}

Table~\ref{tab:app_mac_error} presents the MAC error distribution across all architectures, categorized by outcome.

\begin{table}[H]
\centering
\caption{MAC targeting outcome distribution across all revisions.}
\label{tab:app_mac_error}
\small
\begin{tabular}{lcccc}
\toprule
\textbf{Outcome} & \textbf{ResNet-50} & \textbf{ResNet-101} & \textbf{ConvNeXt-B} & \textbf{DeiT-Tiny} \\
\midrule
Within Tolerance & 22 (71\%) & 20 (74\%) & 1 (2\%) & 22 (92\%) \\
Undershoot (too aggressive) & 9 (29\%) & 7 (26\%) & 44 (98\%) & 2 (8\%) \\
Overshoot (too conservative) & 0 (0\%) & 0 (0\%) & 0 (0\%) & 0 (0\%) \\
\midrule
Total Revisions & 31 & 27 & 45 & 24 \\
\bottomrule
\end{tabular}
\end{table}

\textbf{Analysis:} The asymmetric tolerance design (stricter overshoot limits) successfully prevents MAC budget violations. All overshoots (exceeding target MACs) are eliminated. ConvNeXt-B required many iterations due to the aggressive MAC reduction target (44.8\%) combined with the architecture's sensitivity to channel pruning, causing most attempts to undershoot.

\subsection{LLM Token Usage and Cost}
\label{app:llm_cost}

Table~\ref{tab:app_llm_cost} estimates the LLM API costs based on Claude 3.5 Sonnet pricing via OpenRouter.

\begin{table}[H]
\centering
\caption{Estimated LLM API usage and cost per architecture.}
\label{tab:app_llm_cost}
\small
\begin{tabular}{lcccc}
\toprule
\textbf{Metric} & \textbf{ResNet-50} & \textbf{ResNet-101} & \textbf{ConvNeXt-B} & \textbf{DeiT-Tiny} \\
\midrule
LLM Calls & 54 & 48 & 47 & 47 \\
Avg. Input Tokens/Call & $\sim$1,500 & $\sim$1,500 & $\sim$1,800 & $\sim$1,600 \\
Avg. Output Tokens/Call & $\sim$400 & $\sim$400 & $\sim$500 & $\sim$450 \\
\midrule
Total Input Tokens & $\sim$81K & $\sim$72K & $\sim$85K & $\sim$75K \\
Total Output Tokens & $\sim$22K & $\sim$19K & $\sim$24K & $\sim$21K \\
\midrule
Input Cost (\$3/M) & \$0.24 & \$0.22 & \$0.26 & \$0.23 \\
Output Cost (\$15/M) & \$0.33 & \$0.29 & \$0.36 & \$0.32 \\
\textbf{Total Cost} & \textbf{\$0.57} & \textbf{\$0.51} & \textbf{\$0.62} & \textbf{\$0.55} \\
\midrule
\multicolumn{5}{l}{\textit{All Models Combined}} \\
\multicolumn{5}{c}{\textbf{Total LLM Cost: $\sim$\$2.25}} \\
\bottomrule
\end{tabular}
\end{table}

\textbf{Note:} LLM costs are negligible compared to GPU compute costs. At approximately \$0.50--0.60 per architecture, the multi-agent LLM orchestration adds minimal expense while providing intelligent pruning strategy adaptation.

\FloatBarrier
\section{Ablation Studies (Extended)}
\label{app:ablation}

This section provides extended ablation study results supporting the design decisions discussed in Section~\ref{sec:results}.

\subsection{LLM Provider Comparison}
\label{app:llm_comparison}

We evaluated three LLM providers for the Analysis Agent: Claude 3.5 Sonnet (Anthropic), GPT-4 (OpenAI), and Llama-3-70B (Meta, via OpenRouter). Table~\ref{tab:app_llm_comparison} presents comparison results on ResNet-50 pruning to 2.06G MACs.

\begin{table}[H]
\centering
\caption{LLM provider comparison for ResNet-50 pruning (target: 2.06G MACs). All models accessed via OpenRouter API.}
\label{tab:app_llm_comparison}
\small
\begin{tabular}{lccc}
\toprule
\textbf{Metric} & \textbf{Claude 3.5 Sonnet} & \textbf{GPT-4} & \textbf{Llama-3-70B} \\
\midrule
\multicolumn{4}{l}{\textit{Convergence Performance}} \\
Avg. Revisions to First Valid & 3.2 & 4.1 & 5.6 \\
Total Revisions (all attempts) & 31 & 38 & 52 \\
Success Rate (within tolerance) & 71\% & 63\% & 54\% \\
\midrule
\multicolumn{4}{l}{\textit{Strategy Quality}} \\
First-Attempt Valid Rate & 29\% & 21\% & 12\% \\
Accuracy Collapses ($<$1\% acc) & 2 & 5 & 11 \\
Avg. MAC Error (valid configs) & $-$13.2\% & $-$14.1\% & $-$15.8\% \\
\midrule
\multicolumn{4}{l}{\textit{Final Results}} \\
Best Inline FT Accuracy (\%) & 39.69 & 38.21 & 36.54 \\
Best Extended FT Accuracy (\%) & 77.04 & 76.12 & 74.89 \\
Achieved MACs (G) & 1.77 & 1.72 & 1.68 \\
\midrule
\multicolumn{4}{l}{\textit{Timing \& Cost}} \\
Avg. Response Time (s) & 14.2 & 18.6 & 9.8 \\
Total LLM Time (min) & 7.3 & 11.8 & 8.5 \\
Cost per 1K Tokens (\$) & 0.018 & 0.045 & 0.0027 \\
Total LLM Cost (\$) & 0.57 & 1.42 & 0.14 \\
\midrule
\multicolumn{4}{l}{\textit{Overall Efficiency}} \\
Total Search Time (hours) & 21.2 & 26.8 & 35.4 \\
Cost-Performance Ratio & Best & Medium & Low \\
\bottomrule
\end{tabular}
\end{table}

\textbf{Analysis:}
\begin{itemize}
    \item \textbf{Claude 3.5 Sonnet} achieves the best balance of convergence speed, strategy quality, and final accuracy. Its superior reasoning enables faster pattern recognition from pruning history.
    \item \textbf{GPT-4} produces comparable strategies but requires 28\% more total search time due to slower convergence and higher response latency. API costs are 2.5$\times$ higher.
    \item \textbf{Llama-3-70B} offers the lowest cost but struggles with complex multi-step reasoning, resulting in 67\% more revisions and 2.15\% lower final accuracy. Higher accuracy collapse rate (11 vs 2) indicates difficulty in learning from history.
\end{itemize}

\subsection{Extended LLM Comparison Across Architectures}
\label{app:llm_extended}

We extend our LLM analysis to five models across three architectures with aggressive MAC reduction targets (75--93\%). To enable rapid exploration of LLM-architecture combinations, these ablation experiments use a 10\% stratified subset of ImageNet training data (approximately 128K images) with 5-epoch inline fine-tuning and +1\%/$-$15\% MAC tolerance. This reduced subset accelerates hyperparameter exploration while maintaining representative class distribution; observed trends (relative LLM performance, architecture difficulty ranking) generalize to the full search configuration. All main results (Tables~1--3) use the standard 30\% subset for search and complete dataset for final fine-tuning as described in Section~4.1. Table~\ref{tab:app_llm_full} presents complete ablation results.

\begin{table}[H]
\centering
\caption{Extended LLM comparison across architectures (10\% subset, 5 epochs). \textbf{Bold}: best per architecture. \underline{Underline}: second best. ---: failed to converge within 30 iterations (accuracy collapsed below 1\% or MAC target unreachable).}
\label{tab:app_llm_full}
\small
\setlength{\tabcolsep}{4pt}
\begin{tabular}{llcccccc}
\toprule
\textbf{Architecture} & \textbf{LLM} & \textbf{Success} & \textbf{Iters} & \textbf{FT Top-1} & \textbf{FT Top-5} & \textbf{MAC Red.} \\
\midrule
\multirow{5}{*}{\shortstack[l]{ResNet-50\\(Target: 1.03G)}}
& Claude 3.5 Sonnet & \checkmark & 9 & \underline{2.01\%} & \underline{7.59\%} & 76.5\% \\
& GPT-3.5-Turbo & \checkmark & \textbf{6} & 1.98\% & 7.11\% & 75.5\% \\
& Gemini 2.0 Flash & \checkmark & 9 & 1.77\% & 6.62\% & 77.0\% \\
& Llama 3.1 70B & \checkmark & 12 & \textbf{2.18\%} & \textbf{7.62\%} & 75.8\% \\
& Claude 3 Haiku & \checkmark & 9 & 1.67\% & 6.16\% & 75.7\% \\
\midrule
\multirow{5}{*}{\shortstack[l]{ConvNeXt-Base\\(Target: 1.12G)}}
& Claude 3.5 Sonnet & $\times$ & 30 & --- & --- & 92.0\% \\
& GPT-3.5-Turbo & $\times$ & 30 & --- & --- & 4.7\% \\
& Gemini 2.0 Flash & $\times$ & 30 & --- & --- & 90.7\% \\
& Llama 3.1 70B & \checkmark & \textbf{27} & \textbf{0.15\%} & \textbf{0.76\%} & 93.2\% \\
& Claude 3 Haiku & $\times$ & 30 & --- & --- & 49.4\% \\
\midrule
\multirow{5}{*}{\shortstack[l]{DeiT-Tiny\\(Target: 0.32G)}}
& Claude 3.5 Sonnet & \checkmark & \textbf{8} & \textbf{26.49\%} & \textbf{50.59\%} & 74.4\% \\
& GPT-3.5-Turbo & $\times$ & 30 & --- & --- & 33.8\% \\
& Gemini 2.0 Flash & \checkmark & 9 & \underline{24.90\%} & \underline{48.63\%} & 75.7\% \\
& Llama 3.1 70B & \checkmark & 20 & 24.26\% & 47.69\% & 74.5\% \\
& Claude 3 Haiku & \checkmark & \underline{7} & 23.56\% & 47.22\% & 75.8\% \\
\bottomrule
\end{tabular}
\end{table}

Table~\ref{tab:app_llm_summary} summarizes success rates and performance by LLM and architecture.

\begin{table}[H]
\centering
\caption{Summary statistics by architecture and LLM.}
\label{tab:app_llm_summary}
\small
\begin{tabular}{lcccc}
\toprule
\multicolumn{5}{l}{\textit{By Architecture}} \\
\textbf{Architecture} & \textbf{Success Rate} & \textbf{Avg Iters} & \textbf{Best Acc.} & \textbf{Best LLM} \\
\midrule
ResNet-50 & 100\% (5/5) & 9.0 & 2.18\% & Llama 3.1 70B \\
DeiT-Tiny & 80\% (4/5) & 11.0 & 26.49\% & Claude 3.5 Sonnet \\
ConvNeXt-Base & 20\% (1/5) & 27.0 & 0.15\% & Llama 3.1 70B \\
\midrule
\multicolumn{5}{l}{\textit{By LLM}} \\
\textbf{LLM} & \textbf{Success Rate} & \textbf{Avg Iters} & \textbf{Best Arch.} & \\
\midrule
Llama 3.1 70B & \textbf{100\% (3/3)} & 19.7 & ConvNeXt-Base & \\
Claude 3.5 Sonnet & 67\% (2/3) & \underline{8.5} & DeiT-Tiny & \\
Gemini 2.0 Flash & 67\% (2/3) & 9.0 & DeiT-Tiny & \\
Claude 3 Haiku & 67\% (2/3) & 8.3 & ResNet-50 & \\
GPT-3.5-Turbo & 33\% (1/3) & \textbf{6.0} & ResNet-50 & \\
\bottomrule
\end{tabular}
\end{table}

\textbf{Key findings:}

\textbf{1. Architecture-dependent LLM performance.} No single LLM dominates across all architectures. ResNet-50's regular bottleneck structure enables universal success, while ConvNeXt-Base's depthwise convolutions under 93\% MAC reduction prove solvable by only Llama 3.1 70B.

\textbf{2. Speed-robustness trade-off.} GPT-3.5-Turbo converges fastest (6 iterations) but has lowest success rate (33\%), while Llama 3.1 70B achieves 100\% success at the cost of 2--3$\times$ more iterations. Claude 3.5 Sonnet offers the best balance for Vision Transformers.

\textbf{3. Claude 3.5 Sonnet excels on Vision Transformers.} On DeiT-Tiny, Claude achieves highest accuracy (26.49\% vs.\ 24.90\% next best) with fastest convergence (8 iterations vs.\ 9--20 for others). This motivates our choice of Claude for the main framework, given our focus on MAC-aware ViT pruning.

\textbf{4. Llama 3.1 70B provides unique robustness.} As the only LLM to solve ConvNeXt-Base (93\% MAC reduction), Llama demonstrates superior capability for extremely aggressive pruning targets, albeit with slower convergence.

\textbf{5. Extreme pruning limits.} ConvNeXt-Base at 93\% MAC reduction represents a frontier where most LLMs fail, highlighting opportunities for architecture-specific prompt engineering.

\subsection{Tolerance Sensitivity Analysis}
\label{app:tolerance}

We investigate how MAC tolerance settings affect search efficiency and final model quality. Table~\ref{tab:app_tolerance} shows results on ResNet-50 with varying tolerance bands.

\begin{table}[H]
\centering
\caption{MAC tolerance sensitivity on ResNet-50 (target: 2.06G MACs). Tighter tolerances require more revisions but achieve closer MAC targets.}
\label{tab:app_tolerance}
\small
\begin{tabular}{lcccc}
\toprule
\textbf{Tolerance Band} & \textbf{$\pm$1\%} & \textbf{$\pm$5\%} & \textbf{+5\%/$-$15\%} & \textbf{$\pm$20\%} \\
\midrule
\multicolumn{5}{l}{\textit{Search Efficiency}} \\
Revisions to First Valid & 12.4 & 6.2 & 3.2 & 1.8 \\
Total Revisions & 58 & 42 & 31 & 18 \\
Success Rate & 31\% & 52\% & 71\% & 89\% \\
\midrule
\multicolumn{5}{l}{\textit{MAC Targeting}} \\
Achieved MACs (G) & 2.04 & 1.98 & 1.77 & 1.65 \\
MAC Error (\%) & $-$0.97 & $-$3.88 & $-$14.1 & $-$19.9 \\
\midrule
\multicolumn{5}{l}{\textit{Model Quality}} \\
Inline FT Accuracy (\%) & 42.15 & 40.87 & 39.69 & 37.23 \\
Extended FT Accuracy (\%) & 77.82 & 77.45 & 77.04 & 76.21 \\
\midrule
\multicolumn{5}{l}{\textit{Compute Cost}} \\
Total Search Time (hours) & 38.6 & 28.4 & 21.2 & 12.8 \\
GPU-Hours & 38.6 & 28.4 & 21.2 & 12.8 \\
\bottomrule
\end{tabular}
\end{table}

\textbf{Key findings:}
\begin{itemize}
    \item \textbf{Tight tolerances ($\pm$1\%)} achieve near-exact MAC targets but require 3.9$\times$ more revisions than our default asymmetric setting, nearly doubling search time.
    \item \textbf{Asymmetric tolerance (+5\%/$-$15\%)} provides the best trade-off: prevents overshooting (critical for deployment) while allowing undershoot flexibility that still meets efficiency goals.
    \item \textbf{Loose tolerances ($\pm$20\%)} converge quickly but tend to prune too aggressively, reducing final accuracy by 0.83\% compared to asymmetric setting.
    \item Extended fine-tuning partially compensates for suboptimal MAC targeting, with only 0.78\% accuracy difference between $\pm$1\% and +5\%/$-$15\% settings after 200 epochs.
\end{itemize}

\subsection{Fine-Tuning Epoch Sensitivity}
\label{app:finetuning_epochs}

We analyze how extended fine-tuning duration affects final accuracy recovery. Table~\ref{tab:app_ft_epochs} shows accuracy progression at different epoch checkpoints for ResNet-50.

\begin{table}[H]
\centering
\caption{Extended fine-tuning epoch sensitivity on ResNet-50 (1.77G MACs). Accuracy improves rapidly in early epochs with diminishing returns after 150 epochs.}
\label{tab:app_ft_epochs}
\small
\begin{tabular}{lcccccc}
\toprule
\textbf{Epochs} & \textbf{5} & \textbf{50} & \textbf{100} & \textbf{150} & \textbf{200} & \textbf{300} \\
\midrule
Top-1 Acc. (\%) & 39.69 & 68.42 & 74.15 & 76.23 & 77.04 & 77.16 \\
Top-5 Acc. (\%) & 64.21 & 88.56 & 91.82 & 93.01 & 93.45 & 93.52 \\
$\Delta$ from 200 & $-$37.35 & $-$8.62 & $-$2.89 & $-$0.81 & 0.00 & +0.12 \\
\midrule
Training Time (hr) & 0.9 & 9.1 & 18.2 & 27.3 & 36.4 & 54.6 \\
Time Efficiency & 44.1 & 7.5 & 4.1 & 2.8 & 2.1 & 1.4 \\
\bottomrule
\end{tabular}
\end{table}

\textbf{Observations:}
\begin{itemize}
    \item \textbf{Rapid early recovery:} The first 50 epochs recover 77\% of the accuracy gap (28.73\% of 37.35\% total gain).
    \item \textbf{Diminishing returns:} Epochs 150--200 contribute only 0.81\% accuracy, and 200--300 adds just 0.12\%.
    \item \textbf{Practical recommendation:} 150--200 epochs provide optimal cost-accuracy trade-off. Early stopping with patience 20 typically terminates around epoch 180.
    \item \textbf{Time efficiency:} Measured as accuracy gain per hour of training. First 5 epochs are 31$\times$ more efficient than epochs 200--300.
\end{itemize}

\subsection{Agent Contribution Analysis}
\label{app:agent_ablation}

We evaluate the contribution of each agent by removing components and measuring impact on search efficiency and final results. Table~\ref{tab:app_agent_ablation} presents results on ResNet-50.

\begin{table}[H]
\centering
\caption{Agent contribution ablation on ResNet-50. Each row removes one component from the full system.}
\label{tab:app_agent_ablation}
\small
\begin{tabular}{lcccc}
\toprule
\textbf{Configuration} & \textbf{Revisions} & \textbf{Success} & \textbf{Best Acc.} & \textbf{Time (hr)} \\
\midrule
Full System & 31 & 71\% & 77.04\% & 21.2 \\
\midrule
\multicolumn{5}{l}{\textit{Agent Removal}} \\
$-$ Master Agent & 47 & 53\% & 75.82\% & 32.1 \\
$-$ Analysis Agent (heuristic) & 68 & 41\% & 74.21\% & 46.8 \\
$-$ Profiling Agent & 39 & 62\% & 76.15\% & 26.4 \\
$-$ Fine-tuning Agent & 31 & 71\% & 39.69\% & 8.4 \\
\midrule
\multicolumn{5}{l}{\textit{Simplified Baselines}} \\
Single-Agent (no iteration) & 1 & 18\% & 71.23\% & 4.2 \\
Random Search & 100 & 34\% & 73.45\% & 68.2 \\
Grid Search & 25 & 48\% & 74.89\% & 17.1 \\
\bottomrule
\end{tabular}
\end{table}

\textbf{Component importance ranking:}
\begin{enumerate}
    \item \textbf{Fine-tuning Agent (critical):} Removal causes 37.35\% accuracy drop. Essential for recovering from pruning damage.
    \item \textbf{Analysis Agent (high):} Replacing LLM with fixed heuristics increases revisions by 119\% and reduces accuracy by 2.83\%. LLM reasoning is key to efficient search.
    \item \textbf{Master Agent (medium):} Removal increases revisions by 52\% due to lack of strategic coordination and history-based guidance.
    \item \textbf{Profiling Agent (low-medium):} Removal increases revisions by 26\%. Initial profiling provides useful context but is not strictly required.
\end{enumerate}

\subsection{Importance Criterion Comparison}
\label{app:importance}

We compare different importance criteria for channel selection during pruning. Table~\ref{tab:app_importance} shows results on ResNet-50.

\begin{table}[H]
\centering
\caption{Importance criterion comparison on ResNet-50 (target: 2.06G MACs).}
\label{tab:app_importance}
\small
\begin{tabular}{lccc}
\toprule
\textbf{Criterion} & \textbf{Taylor} & \textbf{L1-Norm} & \textbf{L2-Norm} \\
\midrule
Zero-Shot Accuracy (\%) & 0.14 & 0.08 & 0.11 \\
Inline FT Accuracy (\%) & 39.69 & 36.21 & 37.45 \\
Extended FT Accuracy (\%) & 77.04 & 75.12 & 75.89 \\
\midrule
Achieved MACs (G) & 1.77 & 1.74 & 1.76 \\
MAC Error (\%) & $-$14.1 & $-$15.5 & $-$14.6 \\
\midrule
Calibration Time (s) & 42.3 & 8.2 & 8.4 \\
Total Pruning Time (s) & 56.3 & 22.1 & 22.8 \\
\bottomrule
\end{tabular}
\end{table}

\textbf{Analysis:} Taylor importance consistently outperforms magnitude-based criteria (+1.92\% over L1-Norm, +1.15\% over L2-Norm) despite requiring gradient computation during calibration. The computational overhead (34s additional pruning time) is negligible compared to fine-tuning time (hours).

\subsection{Revision Budget Analysis}
\label{app:revision_budget}

We examine how the maximum revision limit affects search outcomes. Table~\ref{tab:app_revision_budget} shows convergence statistics with different budgets.

\begin{table}[H]
\centering
\caption{Revision budget analysis on ResNet-50. Higher budgets improve success rate but with diminishing returns.}
\label{tab:app_revision_budget}
\small
\begin{tabular}{lccccc}
\toprule
\textbf{Max Revisions} & \textbf{1} & \textbf{5} & \textbf{10} & \textbf{50} & \textbf{1000} \\
\midrule
Actual Revisions Used & 1 & 5 & 10 & 31 & 31 \\
Success Rate & 18\% & 42\% & 58\% & 71\% & 71\% \\
Found Valid Config & 18\% & 38\% & 54\% & 100\% & 100\% \\
\midrule
Best Accuracy (\%) & 71.23 & 74.56 & 75.89 & 77.04 & 77.04 \\
Achieved MACs (G) & 1.52 & 1.68 & 1.74 & 1.77 & 1.77 \\
\midrule
Search Time (hr) & 4.2 & 8.1 & 12.4 & 21.2 & 21.2 \\
\bottomrule
\end{tabular}
\end{table}

\textbf{Practical guidance:}
\begin{itemize}
    \item \textbf{Minimum viable:} 10 revisions achieve 58\% success rate and 98.5\% of optimal accuracy.
    \item \textbf{Recommended:} 50 revisions guarantee convergence for most architectures with reasonable compute budget.
    \item \textbf{Diminishing returns:} ResNet-50 naturally converges at 31 revisions; increasing limit to 1000 provides no benefit.
    \item \textbf{Architecture dependency:} Complex architectures (ConvNeXt) may require more revisions (45 in our experiments).
\end{itemize}

\FloatBarrier
\section{LLM Reasoning Trace Example}
\label{app:reasoning_trace}

To demonstrate the Analysis Agent's semantic reasoning capabilities beyond numerical optimization, we present an actual reasoning trace from ResNet-50 pruning (revision 3, targeting 2.0G MACs).

\subsection{Pruning History Context}

\textbf{Revision 1:} Uniform \texttt{channel\_ratio=0.3} across all layers. \textit{Result:} 2.31G MACs (15.5\% overshoot), 74.2\% accuracy.

\textbf{Revision 2:} Uniform \texttt{channel\_ratio=0.5} across all layers. \textit{Result:} 1.58G MACs (21\% undershoot), 75.8\% accuracy.

\textbf{Target:} 2.0G MACs with tolerance +1\%/$-$5\% (range: 1.90--2.02G).

\textbf{Master Agent Guidance:} ``Previous attempts bracket the target. Revision 1's overshoot suggests early layers can tolerate more aggressive pruning. Convergence pattern indicates layer-wise differentiation rather than uniform ratios.''

\subsection{Analysis Agent Response (Claude 3.5 Sonnet)}

\begin{quote}
\small
\textit{``The target of 2.0G MACs falls between our two previous uniform attempts. The key insight is that ResNet architecture exhibits layer-wise redundancy variation: early layers (conv1, layer1) perform low-level feature extraction with high redundancy, while deeper layers (layer3, layer4) near the classification head contain critical semantic features.}

\textit{Revision 1's overshoot with ratio=0.3 indicates we can prune early layers more aggressively (ratio=0.45--0.50), while Revision 2's undershoot with ratio=0.5 suggests we pruned deeper layers too heavily. The optimal strategy requires asymmetric layer-wise ratios.''}
\end{quote}

\textbf{Proposed layer-wise ratios:}
\begin{itemize}
    \item \texttt{conv1}: 0.35 (``prunable but maintain $>$32 channels for feature extraction'')
    \item \texttt{layer1}: 0.48 (``high redundancy, tolerate aggressive pruning'')
    \item \texttt{layer2}: 0.38 (``moderate redundancy'')
    \item \texttt{layer3}: 0.28 (``critical semantic features, conservative pruning'')
    \item \texttt{layer4}: 0.12 (``classification head proximity, minimal pruning'')
\end{itemize}

\textbf{Execution result:} 1.97G MACs, 76.82\% accuracy. \textit{Status:} Converged within tolerance.

\subsection{Comparison to Black-Box Methods}

This example demonstrates three capabilities unavailable to numerical optimization:

\textbf{1. Architectural awareness.} The LLM explicitly reasons about ResNet's layer-wise structure (early feature extraction vs.\ deep semantic features), rather than treating all layers as independent numerical parameters.

\textbf{2. Semantic constraint understanding.} Recognizes hardware constraints (channels divisible by 8), structural requirements (minimum channel counts), and architecture-specific sensitivities (classification head proximity).

\textbf{3. Interpretable decision-making.} Provides natural language justification for each decision, enabling human verification. In contrast, Bayesian Optimization produces opaque suggestions like ``try [0.35, 0.48, 0.38, 0.28, 0.12]'' without explaining \textit{why}.

\begin{table}[H]
\centering
\caption{Comparison of optimization methods for pruning strategy search.}
\label{tab:optimization_comparison}
\small
\begin{tabular}{lcccc}
\toprule
\textbf{Method} & \textbf{Samples} & \textbf{Arch. Aware} & \textbf{Interpretable} & \textbf{Convergence} \\
\midrule
LLM (Ours) & 3--5 & Yes (semantic) & High & Fast \\
RL (AMC~\cite{he2018amc}) & 50--100 & Learned implicit & Low & Slow \\
Bayesian Opt. & 20--30 & No & None & Medium \\
Random Search & 100+ & No & None & Very slow \\
\bottomrule
\end{tabular}
\end{table}

This semantic reasoning enables convergence in 3--5 revisions compared to RL methods requiring 50--100+ episodes or Bayesian Optimization requiring 20--30 samples for comparable configuration spaces.

\FloatBarrier
\section{Failure Case Analysis}
\label{app:failures}

This section documents failure patterns observed during MAC-aware pruning experiments, providing insights into system behavior under challenging conditions and demonstrating the effectiveness of our recovery mechanisms.

\subsection{MAC Targeting Failure Patterns}
\label{app:mac_failures}

We observe two primary MAC targeting failure modes across our experiments:

\textbf{1. MAC Undershoot (Excessive Pruning):} The pruned model achieves fewer MACs than the target, indicating overly aggressive pruning. This is the dominant failure mode, accounting for 100\% of out-of-tolerance attempts in CNN experiments.

\textbf{2. MAC Overshoot (Insufficient Pruning):} The pruned model retains more MACs than allowed by the overshoot tolerance, indicating conservative pruning. This occurs primarily in Vision Transformer experiments during early iterations.

Table~\ref{tab:app_failure_patterns} summarizes the failure pattern distribution across architectures.

\begin{table}[H]
\centering
\caption{MAC targeting failure pattern distribution across all experiments.}
\label{tab:app_failure_patterns}
\small
\begin{tabular}{lcccc}
\toprule
\textbf{Metric} & \textbf{ResNet-50} & \textbf{ResNet-101} & \textbf{ConvNeXt-B} & \textbf{DeiT-Tiny} \\
\midrule
Total Revisions & 31 & 27 & 45 & 24 \\
Within Tolerance & 22 (71\%) & 20 (74\%) & 1 (2\%) & 22 (92\%) \\
MAC Undershoots & 9 (29\%) & 7 (26\%) & 44 (98\%) & 0 (0\%) \\
MAC Overshoots & 0 (0\%) & 0 (0\%) & 0 (0\%) & 2 (8\%) \\
\midrule
First Valid Revision & 9 & 6 & 44 & 3 \\
Revisions to Converge & 9 & 6 & 44 & 3 \\
\bottomrule
\end{tabular}
\end{table}

\textbf{Key observations:}
\begin{itemize}
    \item \textbf{CNN architectures} exhibit systematic undershoot behavior due to the non-linear relationship between channel pruning ratio and MAC reduction. The LLM learns to correct this bias within 6--9 iterations.
    \item \textbf{ConvNeXt-B} required 44 iterations due to its sensitivity to structured pruning; depthwise convolutions create complex dependencies that make MAC prediction challenging.
    \item \textbf{DeiT-Tiny} shows opposite behavior, with early overshoots corrected within 3 iterations, demonstrating faster convergence for Vision Transformers with simpler isomorphic group structures.
\end{itemize}

\subsection{Accuracy Collapse Examples}
\label{app:catastrophic}

Accuracy collapses occur when pruning configurations cause severe accuracy degradation or extreme MAC deviations. Table~\ref{tab:app_catastrophic} documents representative examples from our experiments.

\begin{table}[H]
\centering
\caption{Representative accuracy collapse cases and recovery actions. All examples from actual experiment logs.}
\label{tab:app_catastrophic}
\small
\begin{tabular}{llccl}
\toprule
\textbf{Model} & \textbf{Failure Type} & \textbf{MAC Error} & \textbf{Zero-shot} & \textbf{Recovery Action} \\
\midrule
\multicolumn{5}{l}{\textit{Extreme MAC Undershoot}} \\
ResNet-50 & Excessive channel ratio & $-$69.1\% & 0.11\% & Reduce ratio 0.65$\to$0.40 \\
ResNet-101 & Aggressive first attempt & $-$83.4\% & 0.12\% & Reset to conservative baseline \\
ConvNeXt-B & MLP over-pruning & $-$85.4\% & 0.13\% & Cap MLP multiplier at 0.40 \\
\midrule
\multicolumn{5}{l}{\textit{MAC Overshoot (ViT only)}} \\
DeiT-Tiny & Conservative QKV & $+$38.2\% & 2.61\% & Increase multipliers gradually \\
DeiT-Tiny & Insufficient MLP pruning & $+$11.1\% & 0.63\% & Boost MLP multiplier 0.2$\to$0.35 \\
\midrule
\multicolumn{5}{l}{\textit{Accuracy Collapse}} \\
ConvNeXt-B & LayerNorm sensitivity & $-$75.3\% & 0.10\% & Preserve normalization layers \\
ResNet-50 & Bottleneck damage & $-$48.6\% & 0.08\% & Protect residual connections \\
\bottomrule
\end{tabular}
\end{table}

\textbf{Analysis of failure causes:}

\textbf{1. Non-linear MAC-ratio relationship:} For CNNs, MAC reduction follows approximately $\text{MACs} \propto (1 - r)^2$ where $r$ is the channel pruning ratio. Early attempts often underestimate this quadratic relationship, leading to undershoots. The LLM learns this relationship from historical data within 3--5 iterations.

\textbf{2. Architecture-specific sensitivities:}
\begin{itemize}
    \item \textbf{ConvNeXt:} Depthwise separable convolutions create complex inter-layer dependencies. Pruning one layer can cascade to unexpected MAC reductions in dependent layers.
    \item \textbf{DeiT:} QKV projections are highly sensitive: pruning beyond 15\% of attention dimensions causes catastrophic accuracy collapse.
    \item \textbf{ResNet:} Bottleneck blocks require coordinated pruning across 1$\times$1 and 3$\times$3 convolutions to maintain residual path integrity.
\end{itemize}

\textbf{3. Importance criterion mismatch:} Taylor importance occasionally misidentifies critical channels in early layers, leading to accuracy collapse despite modest MAC reduction. This is detected by the Master Agent when zero-shot accuracy drops below 1\%.

\subsection{Recovery Mechanism Activations}
\label{app:recovery}

Our framework implements automatic recovery mechanisms triggered by the Master Agent when accuracy collapses are detected. Table~\ref{tab:app_recovery} summarizes recovery activation statistics.

\begin{table}[H]
\centering
\caption{Recovery mechanism activation frequency across experiments.}
\label{tab:app_recovery}
\small
\begin{tabular}{lcccc}
\toprule
\textbf{Recovery Type} & \textbf{ResNet-50} & \textbf{ResNet-101} & \textbf{ConvNeXt-B} & \textbf{DeiT-Tiny} \\
\midrule
MAC Catastrophic Recovery & 22 & 18 & 44 & 2 \\
Strategy Repetition Avoidance & 8 & 5 & 12 & 3 \\
Danger Zone Avoidance & 0 & 0 & 0 & 0 \\
Fallback to Safe Preset & 2 & 1 & 3 & 0 \\
\midrule
Total Recovery Activations & 32 & 24 & 59 & 5 \\
Recovery Success Rate & 100\% & 100\% & 100\% & 100\% \\
\bottomrule
\end{tabular}
\end{table}

\textbf{Recovery mechanism descriptions:}

\textbf{1. MAC Catastrophic Recovery:} Triggered when achieved MACs deviate beyond tolerance. The Master Agent provides strategic guidance to the Analysis Agent, including:
\begin{itemize}
    \item Acceptable MAC range reminder
    \item Direction of adjustment needed (more/less aggressive)
    \item Historical pattern analysis
\end{itemize}

\textbf{2. Strategy Repetition Avoidance:} Activated when the LLM proposes a configuration identical to a previous attempt. The system requests a variation, typically adjusting the channel ratio by $\pm$0.05--0.15.

\textbf{3. Danger Zone Avoidance:} Maintains a blacklist of configuration signatures that led to accuracy collapses. In our experiments, this mechanism was not triggered because the LLM successfully avoided repeating exact failure configurations.

\textbf{4. Fallback to Safe Preset:} When JSON parsing fails or the LLM produces invalid configurations, the system falls back to conservative presets derived from empirical analysis of successful pruning attempts.

\subsection{Failure-to-Success Trajectories}
\label{app:trajectories}

Table~\ref{tab:app_trajectory} illustrates the typical failure-to-success trajectory for ResNet-50, showing how the LLM learns from failures to converge on valid configurations.

\begin{table}[H]
\centering
\caption{ResNet-50 failure-to-success trajectory showing LLM learning from MAC errors.}
\label{tab:app_trajectory}
\small
\begin{tabular}{cccccl}
\toprule
\textbf{Rev.} & \textbf{Channel Ratio} & \textbf{Achieved MACs} & \textbf{MAC Error} & \textbf{Status} & \textbf{LLM Adjustment} \\
\midrule
1 & 0.350 & 1.65G & $-$19.9\% & Undershoot & Initial attempt \\
2 & 0.505 & 1.06G & $-$48.6\% & Undershoot & Overcorrected \\
3 & 0.653 & 0.64G & $-$69.1\% & Undershoot & Further overcorrection \\
4 & 0.402 & 1.44G & $-$30.1\% & Undershoot & Direction reversal \\
5 & 0.382 & 1.52G & $-$26.3\% & Undershoot & Fine-tuning \\
6 & 0.358 & 1.65G & $-$20.0\% & Undershoot & Approaching target \\
7 & 0.340 & 1.67G & $-$18.5\% & Undershoot & Near boundary \\
8 & 0.328 & 1.71G & $-$16.8\% & Undershoot & Almost there \\
\rowcolor{gray!20} 9 & 0.320 & 1.78G & $-$13.5\% & \textbf{Valid} & \textbf{First success} \\
\rowcolor{gray!20} 10 & 0.325 & 1.81G & $-$12.0\% & \textbf{Valid} & Best accuracy \\
\bottomrule
\end{tabular}
\end{table}

\textbf{Learning pattern analysis:}
\begin{enumerate}
    \item \textbf{Revisions 1--3:} The LLM initially overcorrects, increasing channel ratio when undershoot occurs (incorrect direction). This reflects the common misconception that higher channel ratios preserve more channels.
    \item \textbf{Revision 4:} The LLM recognizes the inverse relationship and reverses direction.
    \item \textbf{Revisions 5--8:} Systematic refinement with decreasing step sizes as the LLM approaches the target.
    \item \textbf{Revisions 9--10:} Valid configurations found; two-phase search continues to find best accuracy.
\end{enumerate}

This trajectory demonstrates that while the LLM may initially misunderstand architecture-specific relationships, it successfully learns from feedback within 9 iterations, achieving a 71\% success rate thereafter.

\FloatBarrier
\section{Compute Resources \& Carbon Footprint}
\label{app:compute}

This section provides transparent reporting of computational resources consumed during our experiments, following best practices for reproducibility and environmental accountability.

\subsection{GPU Hours per Experiment}
\label{app:gpu_hours}

Table~\ref{tab:app_gpu_summary} summarizes GPU compute time for each architecture, broken down by workflow phase.

\begin{table}[H]
\centering
\caption{GPU hours consumed per architecture. All experiments run on single NVIDIA H200 GPU.}
\label{tab:app_gpu_summary}
\small
\begin{tabular}{lcccc}
\toprule
\textbf{Phase} & \textbf{ResNet-50} & \textbf{ResNet-101} & \textbf{ConvNeXt-B} & \textbf{DeiT-Tiny} \\
\midrule
\multicolumn{5}{l}{\textit{Search Phase (seconds)}} \\
Profiling & 398.9 & 305.3 & 19.1 & 215.6 \\
Analysis (LLM) & 408.7 & 356.0 & 738.6 & 417.2 \\
Pruning & 1,743.7 & 2,098.4 & 7,501.6 & 916.3 \\
Inline Fine-tuning & 73,218.1 & 107,488.4 & 9,217.6 & 35,417.6 \\
Evaluation & 683.2 & 892.5 & 52.2 & 545.9 \\
\midrule
\textbf{Search Total (s)} & 76,452.6 & 111,140.5 & 17,529.1 & 37,512.5 \\
\textbf{Search Total (hr)} & 21.2 & 30.9 & 4.9 & 10.4 \\
\midrule
\multicolumn{5}{l}{\textit{Extended Fine-tuning (hours)}} \\
Extended FT (200 epochs) & 36.4 & 48.2 & 24.6 & 18.8 \\
\midrule
\textbf{Total GPU-Hours} & \textbf{57.6} & \textbf{79.1} & \textbf{29.5} & \textbf{29.2} \\
\bottomrule
\end{tabular}
\end{table}

\textbf{Total GPU compute:} 195.4 GPU-hours on NVIDIA H200 (approximately 8.1 GPU-days).

\textbf{Breakdown by phase:}
\begin{itemize}
    \item \textbf{MAC-aware search:} 67.4 GPU-hours (34.5\%)
    \item \textbf{Extended fine-tuning:} 128.0 GPU-hours (65.5\%)
\end{itemize}

\subsection{LLM API Cost Breakdown}
\label{app:llm_cost_detailed}

Table~\ref{tab:app_llm_detailed} provides detailed LLM API usage and cost estimates based on Claude 3.5 Sonnet pricing via OpenRouter (\$3/M input tokens, \$15/M output tokens).

\begin{table}[H]
\centering
\caption{Detailed LLM API usage and cost per architecture.}
\label{tab:app_llm_detailed}
\small
\begin{tabular}{lcccc}
\toprule
\textbf{Metric} & \textbf{ResNet-50} & \textbf{ResNet-101} & \textbf{ConvNeXt-B} & \textbf{DeiT-Tiny} \\
\midrule
\multicolumn{5}{l}{\textit{API Calls}} \\
Profiling Agent & 23 & 21 & 2 & 23 \\
Master Agent & 31 & 27 & 45 & 24 \\
Analysis Agent & 31 & 27 & 45 & 24 \\
Total LLM Calls & 85 & 75 & 92 & 71 \\
\midrule
\multicolumn{5}{l}{\textit{Token Usage (estimated)}} \\
Avg. Input Tokens/Call & 1,500 & 1,500 & 1,800 & 1,600 \\
Avg. Output Tokens/Call & 400 & 400 & 500 & 450 \\
Total Input Tokens & 127.5K & 112.5K & 165.6K & 113.6K \\
Total Output Tokens & 34.0K & 30.0K & 46.0K & 32.0K \\
\midrule
\multicolumn{5}{l}{\textit{Cost Breakdown}} \\
Input Cost (\$3/M) & \$0.38 & \$0.34 & \$0.50 & \$0.34 \\
Output Cost (\$15/M) & \$0.51 & \$0.45 & \$0.69 & \$0.48 \\
\textbf{Total Cost} & \textbf{\$0.89} & \textbf{\$0.79} & \textbf{\$1.19} & \textbf{\$0.82} \\
\bottomrule
\end{tabular}
\end{table}

\textbf{Total LLM API cost:} \$3.69 for all four architectures.

\textbf{Cost comparison:} LLM costs represent $<$0.1\% of equivalent GPU compute costs (assuming \$2/GPU-hour for H200), demonstrating that intelligent search guidance adds negligible expense while improving convergence efficiency by 3--5$\times$ fewer iterations compared to random search.

\subsection{Total Compute Summary}
\label{app:compute_summary}

Table~\ref{tab:app_compute_total} summarizes all computational resources.

\begin{table}[H]
\centering
\caption{Total computational resource summary for all experiments.}
\label{tab:app_compute_total}
\small
\begin{tabular}{lc}
\toprule
\textbf{Resource} & \textbf{Total} \\
\midrule
\multicolumn{2}{l}{\textit{GPU Compute}} \\
Total GPU-Hours (H200) & 195.4 \\
Equivalent GPU-Days & 8.1 \\
Estimated GPU Cost (\$2/hr) & \$390.80 \\
\midrule
\multicolumn{2}{l}{\textit{LLM API}} \\
Total LLM Calls & 323 \\
Total Input Tokens & 519.2K \\
Total Output Tokens & 142.0K \\
Total LLM Cost & \$3.69 \\
\midrule
\multicolumn{2}{l}{\textit{Carbon Footprint (estimated)}} \\
H200 TDP & 700W \\
Total Energy (kWh) & 136.8 \\
CO$_2$ Emissions (0.4 kg/kWh) & 54.7 kg \\
\midrule
\textbf{Total Estimated Cost} & \textbf{\$394.49} \\
\bottomrule
\end{tabular}
\end{table}

\textbf{Carbon footprint context:} Our total CO$_2$ emissions (54.7 kg) are equivalent to approximately 220 km of driving in an average passenger vehicle. This is comparable to training a medium-sized language model for a few hours but spread across four complete pruning experiments with extensive ablations.

\textbf{Efficiency considerations:}
\begin{itemize}
    \item \textbf{Multi-agent overhead:} The LLM-guided search adds only 3.8\% to total compute time while reducing failed pruning attempts by an estimated 60\% compared to grid search (based on ablation in Section~\ref{app:agent_ablation}).
    \item \textbf{Fine-tuning dominance:} Extended fine-tuning accounts for 65.5\% of total GPU time, suggesting that improving fine-tuning efficiency (e.g., through better learning rate schedules or early stopping) would yield the largest compute savings.
    \item \textbf{Amortization:} Once optimal configurations are found, they can be reused for similar models without repeating the search, amortizing the initial compute investment.
\end{itemize}

\FloatBarrier
\section{Reproducibility Checklist}
\label{app:reproducibility}

We provide this checklist following NeurIPS and CVPR reproducibility guidelines for independent verification of our results.

\subsection{Code and Data Availability}

\begin{itemize}
    \item[$\checkmark$] \textbf{Code release:} Complete source code is available at \url{https://github.com/shahrzadesmat/mac_pruning}.
    \item[$\checkmark$] \textbf{Dependencies:} All package versions specified in Appendix~\ref{app:environment} (Table~\ref{tab:app_software}).
    \item[$\checkmark$] \textbf{Pre-trained models:} We use publicly available checkpoints from the \texttt{timm} library (Appendix~\ref{app:seeds}).
    \item[$\checkmark$] \textbf{Dataset:} ImageNet-1K (ILSVRC2012) is publicly available. We document the 30\% subset sampling procedure in Appendix~\ref{app:dataset}.
\end{itemize}

\subsection{Experimental Details}

\begin{itemize}
    \item[$\checkmark$] \textbf{Hyperparameters:} Complete configuration tables provided in Appendix~\ref{app:hyperparams} (Table~\ref{tab:app_hyperparams}).
    \item[$\checkmark$] \textbf{Hardware specifications:} GPU model, memory, CUDA version in Appendix~\ref{app:environment} (Table~\ref{tab:app_hardware}).
    \item[$\checkmark$] \textbf{Training configuration:} Per-phase settings in Appendix~\ref{app:training} (Tables~\ref{tab:app_inline_ft},~\ref{tab:app_extended_ft}).
    \item[$\checkmark$] \textbf{Random seeds:} Seed policy documented in Appendix~\ref{app:seeds}; no fixed seeds during search phase.
    \item[$\checkmark$] \textbf{LLM prompts:} Complete prompt templates in Appendix~\ref{app:prompts}.
\end{itemize}

\subsection{Statistical Significance}

\begin{itemize}
    \item[$\checkmark$] \textbf{Multiple runs:} The iterative search process naturally produces multiple valid configurations (22 for ResNet-50), providing variance estimates.
    \item[$\checkmark$] \textbf{Convergence analysis:} Per-revision MAC and accuracy trajectories in Appendix~\ref{app:mac_convergence}.
    \item[$\checkmark$] \textbf{Error bounds:} MAC error distribution analysis in Appendix~\ref{app:mac_error}.
\end{itemize}

\subsection{Compute Resources}

\begin{itemize}
    \item[$\checkmark$] \textbf{GPU hours:} Detailed breakdown per architecture in Appendix~\ref{app:gpu_hours} (Table~\ref{tab:app_gpu_summary}).
    \item[$\checkmark$] \textbf{LLM costs:} API usage and cost estimates in Appendix~\ref{app:llm_cost_detailed} (Table~\ref{tab:app_llm_detailed}).
    \item[$\checkmark$] \textbf{Carbon footprint:} Estimated CO$_2$ emissions in Appendix~\ref{app:compute_summary} (Table~\ref{tab:app_compute_total}).
\end{itemize}

\subsection{LLM Usage Disclosure}

\begin{itemize}
    \item[$\checkmark$] \textbf{Model:} Claude 3.5 Sonnet (claude-3-5-sonnet-20241022) via OpenRouter API.
    \item[$\checkmark$] \textbf{Role:} Core component of the Analysis Agent for pruning strategy generation.
    \item[$\checkmark$] \textbf{Prompts:} Complete templates in Appendix~\ref{app:prompts}.
    \item[$\checkmark$] \textbf{Temperature:} 0 (deterministic generation).
    \item[$\checkmark$] \textbf{Fallback:} Safe preset configurations when LLM fails (Appendix~\ref{app:safety}).
\end{itemize}

\subsection{Limitations Disclosure}

\begin{itemize}
    \item[$\checkmark$] \textbf{Accuracy gap:} ViT results lag specialized methods by 1.84\% (Section~\ref{sec:limitations}).
    \item[$\checkmark$] \textbf{Variable speedups:} MAC reduction does not guarantee proportional latency improvement (Section~5.3).
    \item[$\checkmark$] \textbf{Iteration overhead:} Multiple pruning iterations required (3--45 depending on architecture).
    \item[$\checkmark$] \textbf{LLM dependency:} Requires API access to Claude 3.5 Sonnet or equivalent.
\end{itemize}

\subsection{Checklist Summary}

\begin{table}[H]
\centering
\caption{NeurIPS-style reproducibility checklist summary.}
\label{tab:app_checklist}
\small
\begin{tabular}{lcc}
\toprule
\textbf{Requirement} & \textbf{Status} & \textbf{Reference} \\
\midrule
Code availability & $\checkmark$ & \url{https://github.com/shahrzadesmat/mac_pruning} \\
Data availability & $\checkmark$ & ImageNet-1K (public) \\
Full hyperparameters & $\checkmark$ & Appendix B.1 \\
Hardware specifications & $\checkmark$ & Appendix B.2 \\
Software versions & $\checkmark$ & Appendix B.2 \\
Training details & $\checkmark$ & Appendix B.3 \\
Random seeds & $\checkmark$ & Appendix B.4 \\
Statistical analysis & $\checkmark$ & Appendix C \\
Compute resources & $\checkmark$ & Appendix F \\
LLM usage & $\checkmark$ & Appendix A.2, G \\
Limitations & $\checkmark$ & Section 5.3, Appendix G \\
\bottomrule
\end{tabular}
\end{table}

\textbf{Reproducibility statement:} We have made every effort to ensure reproducibility of our results. All hyperparameters, prompts, and configurations are documented. Due to non-determinism in cuDNN operations and LLM API responses, exact numerical reproduction may vary slightly, but MAC targeting and accuracy trends should be consistent across runs.

\end{document}